\documentclass[11pt]{article}

\usepackage[final]{acl}

\usepackage{times}
\usepackage{latexsym}

\usepackage[T1]{fontenc}

\usepackage[utf8]{inputenc}

\usepackage{microtype}

\usepackage{inconsolata}

\usepackage{graphicx}
\usepackage{amsmath}
\usepackage{booktabs}
\usepackage{array}
\usepackage{multicol}
\usepackage{multirow}
  \usepackage[most]{tcolorbox}
  \usepackage{needspace}      
  \usepackage{enumitem}       

\usepackage[table]{xcolor}

\usepackage{hyperref}

\usepackage{listings}
\usepackage{xcolor}

\title{
  \raisebox{-0.3em}{\includegraphics[height=1.4em]{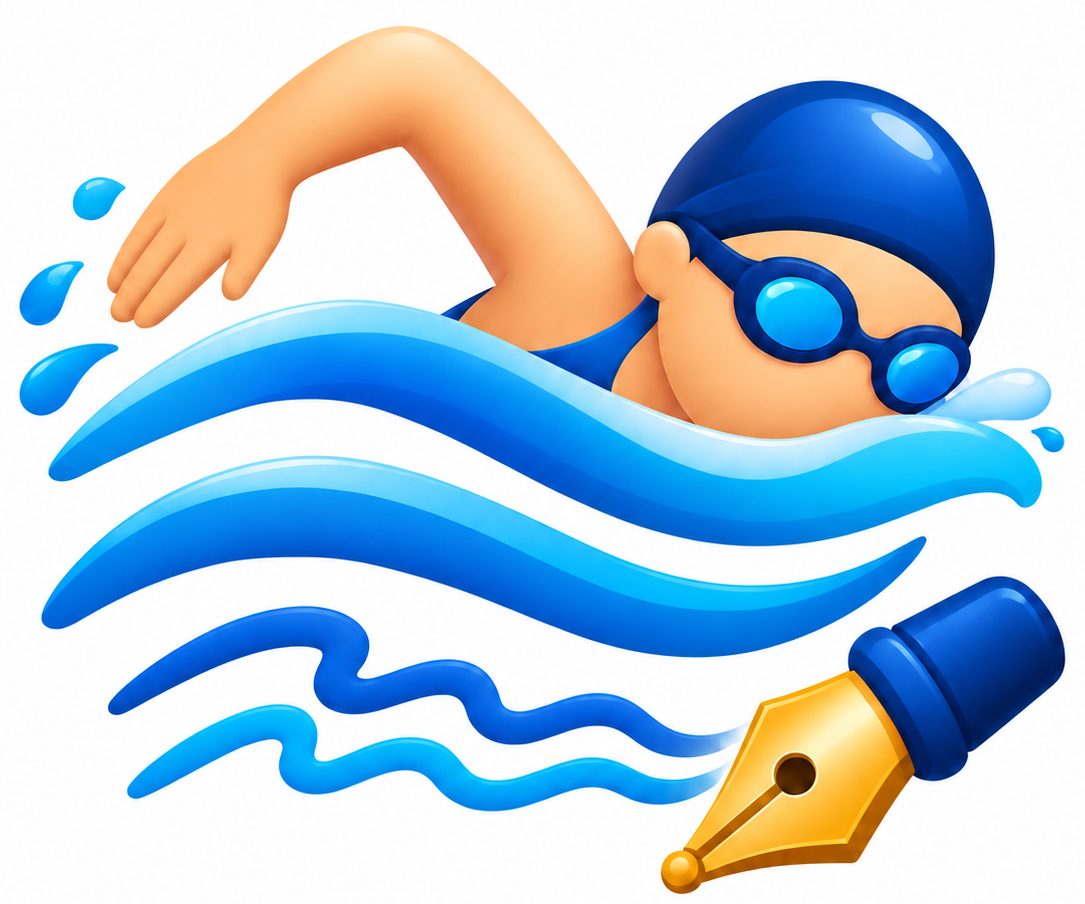}}
  \hspace{0.2em}
  SWIM: Student Writing Simulation\\ via Proficiency-Conditioned Generation
}

\author{
    Heejin Do$^{1, 2}$ \quad
    Jakub Kontak$^{1}$ \quad
    {Mrinmaya Sachan$^{1}$} \\ \text{} \\
  $^{1}$Department of Computer Science, ETH Zurich 
  $^2$ ETH AI Center \\
 \texttt{heejin.do@ai.ethz.ch} \\
 \url{https://github.com/doheejin/SWIM}
}

\usepackage{multirow}
\usepackage{amsmath} 
\usepackage{hyperref}       
\usepackage{cleveref}
\usepackage{url}            
\usepackage{booktabs}       
\usepackage{graphicx}
\usepackage{tabularx}
\usepackage{enumitem}
\usepackage{bm}
\usepackage{xspace}
\usepackage{amssymb}
\usepackage{placeins}

\usepackage{algorithm}
\usepackage{algpseudocode}

\crefname{figure}{Fig.}{Figs.}
\crefname{table}{Table}{Tables}
\crefname{appendix}{App.}{Apps.}
\crefname{section}{\S}{\S\S}
\crefformat{section}{\S#2#1#3} 
\crefname{equation}{Eq.}{Eqs.}
\crefname{algorithm}{Alg.}{Algs.}
\crefname{algocf}{Alg.}{Algs.}
\crefname{defin}{Def.}{Defs.}
\crefname{theorem}{Thm.}{Thms.}
\crefname{lemma}{Lemma}{Lemmas}

\usepackage[textsize=tiny, disable]{todonotes}
\newlist{dialogue}{description}{1}
\setlist[dialogue]{
    labelwidth=1.5cm,
    labelindent=0cm,
    leftmargin=1.8cm,
    labelsep=0.3cm,
    align=left,
    noitemsep,
    topsep=0pt
}

\definecolor{colInter}{HTML}{1F77B4}    
\definecolor{colCorr}{HTML}{2CA02C}   
\definecolor{colErrDesc}{HTML}{FF7F0E}  
\definecolor{colInst}{HTML}{D62728}     
\definecolor{colPlaus}{HTML}{9467BD}  
\definecolor{colCur}{HTML}{17BECF}    

\newif\ifarxiv
\arxivtrue      

\makeatletter
\newcommand{\oset}[3][0.23ex]{%
  \mathrel{\mathop{#3}\limits^{
    \vbox to#1{\kern-2\ex@
    \hbox{$\scriptstyle#2$}\vss}}}}
\makeatother

\begin{document}
\maketitle

\begin{abstract}

Writing proficiency manifests in how students develop content, organize ideas, choose words, and use language. Despite growing interest in LLM-based student simulation, whether LLMs can reproduce such multidimensional variation in extended writing remains largely unexplored. In this work, we explore if language models can realistically simulate \emph{student writing}, and introduce {SWIM}, a task that formulates Student Writing sIMulation as proficiency-conditioned essay generation. We evaluate prompting, supervised fine-tuning (SFT), and reinforcement learning (RL) methods for writing simulation using automated essay scoring as a measure of profile alignment. Extensive experiments reveal that prompting provides limited proficiency control, even for strong proprietary LLMs with rubric-grounded strategies. In particular, while models can adjust content-oriented traits, they struggle to reproduce the lexical, grammatical, and organizational variation in different proficiency levels. SFT substantially improves alignment, while RL with the proposed proficiency-alignment reward yields further gains across all writing traits and essay prompts. Our findings suggest that explicit supervision enables substantially stronger profile alignment than prompting alone, while authentic low-proficiency writing remains challenging to reproduce.

\end{abstract}

\section{Introduction}

Educational AI systems increasingly rely on simulated learners to train tutors, optimize feedback strategies, and evaluate pedagogical interventions without costly human studies \citep{mannekote2025can,nair2024closing,xu2023simulate,lu2024generative}. Recent work has shown that language-model-based student simulators can support closed-loop training pipelines, enabling educational agents to improve through interactions with synthetic learners rather than real students \citep{nair2024closing}.

Despite this growing interest, existing work has focused primarily on short learner responses, tutoring interactions, or action prediction \citep{mannekote2025can,lu2024generative,benedetto2024llmexam}. Writing, however, remains largely unexplored as a student simulation problem. This omission is notable because writing is one of the richest manifestations of student behavior, revealing differences in content development, organization, lexical choice, sentence fluency, and language proficiency \citep{bereiter2013psychology,gregg2016cognitive}.

Modeling student writing is also fundamentally challenging. Writing proficiency is inherently multi-dimensional: two students with the same overall proficiency may exhibit very different strengths and weaknesses across traits such as content, organization, and writing conventions \citep{mathias2018asap++,do2024arts}. A writing simulator must therefore not only produce coherent essays but also control for multiple proficiency-related characteristics simultaneously. Prior work on controllable text generation has shown that satisfying multiple text attributes at once remains difficult, with performance often degrading as constraints accumulate \citep{zhang2023macsum,liu2024benchmarking,urlana2024controllable}, raising the question of whether current models can realistically reproduce proficiency-dependent writing behaviors.

In this work, we investigate the question: \emph{Can language models simulate student writing across proficiency levels?} We introduce SWIM (\textbf{S}tudent \textbf{W}riting s\textbf{IM}ulation), a task formulated as trait-proficiency-conditioned essay generation: given a writing prompt and a target proficiency profile, a model must generate an essay that reflects the specified writing aspects. We study three approaches to proficiency control: (i) rubric-grounded prompting with two grounding strategies, i.e., contrastive trait descriptions motivated by \citep{mannekote2025can} and score-specific rubric lookup extended from \citep{imperial2024standardize}; (ii) supervised fine-tuning (SFT) on real (essay, profile) pairs; and (iii) group relative policy optimization (GRPO) \citep{shao2024deepseekmath} with a \emph{Proficiency Alignment Reward} (PAR), which leverages a frozen multi-trait automated essay scoring (AES) model \citep{do2024arts} to reward agreement with the target profile. 

Experiments on ASAP/ASAP++~\citep{asap-aes,mathias2018asap++} show that explicit training substantially strengthens proficiency control, yet does not fully resolve behavioral realism. Even strong proprietary LLMs with rubric grounding and few-shot demonstrations provide limited and uneven control, aligning far better on content-oriented traits such as \textit{Content}, \textit{Prompt Adherence}, and \textit{Narrativity} than on lexical, grammatical, and organizational dimensions. Direct supervision from authentic score--essay pairs proves far more effective than rubric descriptions alone, enabling open-source 7B and 4B models to move from near-random alignment under prompting to meaningful proficiency control. GRPO with the proposed PAR further improves alignment across every writing trait and essay prompt, producing more balanced profile-level generation. Despite these gains, corpus-level analyses show that models reproduce high-proficiency writing more faithfully than low-proficiency writing: prompting tends to rely on superficial corruption, whereas trained models better recover linguistic structure but retain overly polished surface realization. Additional analyses of memorization and within-profile diversity indicate that these gains reflect learned proficiency-conditioned generation rather than training-set copying or exact-output collapse. Our main contributions are:
\begin{enumerate}
    \item We introduce {SWIM}, a task that formalizes student writing simulation as proficiency-conditioned essay generation, together with an AES-based evaluation framework for measuring profile alignment.
    
    \item We systematically compare rubric-grounded prompting, SFT, and RL across proprietary and open-source LLMs,     revealing a clear hierarchy in their ability to control writing proficiency.
    
    \item We introduce the {Proficiency Alignment Reward (PAR)}, a dense AES-derived reward that directly optimizes profile-level alignment and consistently improves simulation fidelity across writing traits and essay prompts.
    
    \item We characterize the remaining gap between profile alignment and behavioral realism, identifying low-proficiency linguistic form as a persistent bottleneck and distinguishing the failure modes of different approaches.
\end{enumerate}

\section{Related Work}

\paragraph{LLM-based Student Simulation.}

Student simulation has long been studied in intelligent tutoring systems for modeling learner behavior, evaluating pedagogical strategies, and developing adaptive educational systems~\cite{pavlik2013review,kass1989student}. Recent advances in LLMs have renewed interest in this area, enabling the simulation of learner responses in tutoring dialogues, interactive learning environments, and agent-training pipelines~\cite{mannekote2025can,xu2023simulate,lu2024generative,nair2024closing,benedetto2024llmexam,scarlatos-etal-2025-smart}. Such simulators provide a scalable alternative to costly human studies and have been used to train educational agents, generate synthetic learner interactions, and evaluate instructional interventions~\cite{nair2024closing,lu2024generative}. Despite this progress, existing work has focused primarily on short responses, dialogue interactions, or learner actions~\cite{mannekote2025can,lu2024generative,benedetto2024llmexam}. Writing remains largely unexplored, despite being representative student behavior and reflecting multiple dimensions of proficiency. In contrast, we extend student simulation to long-form writing and investigate whether LLMs can generate essays aligned with specified proficiency profiles.

\paragraph{Writing Assessment and Student Writing Modeling.}

Educational writing research has largely focused on automated essay scoring (AES), writing evaluation, and feedback generation~\cite{uto2021review,dong2017attention,attali2006erater,ramineni2012evaluation,li2024aes,lee-etal-2024-unleashing,do2024autoregressive_rl}. Recent neural and LLM-based AES systems achieve strong agreement with human raters at both holistic and trait levels~\cite{wang-etal-2022-use,li2024aes,chu-etal-2025-rationale,DO2026132119}. In contrast, relatively little work has investigated the inverse problem of modeling how students write. We formulate student writing simulation as a generative task and leverage AES models as scalable evaluators of simulation fidelity.

\begin{figure*}
    \centering
    \includegraphics[width=0.96\linewidth]{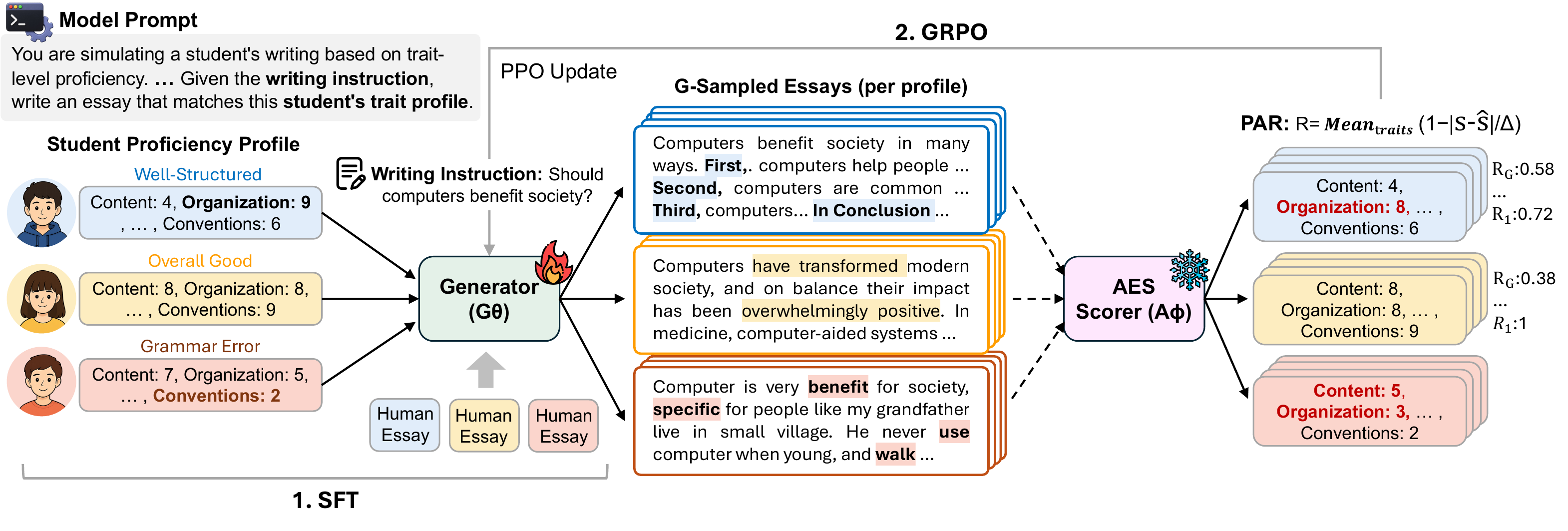}
   \caption{Two-stage training for SWIM. \textbf{1. SFT:} the Generator is first fine-tuned on real (essay, profile) pairs via cross-entropy. \textbf{2. GRPO:} for each profile in a batch, the Generator samples $G$ essays (stacked layers) that are independently scored by the frozen AES Scorer to yield per-essay PAR values $\{R_j\}_{j=1}^{G}$; group-normalized advantages within each profile's group drive the PPO update. The complete prompts are provided in Appendix~\ref{append:prompt}.
}
\label{fig:main_fig}
\end{figure*}

\paragraph{Controllable Text Generation.}

Controllable text generation aims to steer language models toward desired attributes such as topic, style, length, readability, or factuality~\cite{liang2024ctgsurvey,zhang2023macsum,urlana2024controllable}. Prior work has explored a range of prompting- and training-based approaches, including instruction prompting, retrieval augmentation, supervised fine-tuning, and reinforcement learning~\cite{li2024tole,imperial2024standardize,ryu2024iterative}. Despite substantial progress, controlling multiple attributes simultaneously remains challenging, with performance often degrading as the number of constraints increases~\cite{zhang2023macsum,liu2024benchmarking,urlana2024controllable,ryu2024iterative,li2024tole}. 
Several studies have incorporated expert-defined standards or rubrics to improve controllability in educational settings. For example, STANDARDIZE~\cite{imperial2024standardize} uses retrieval-based prompting to ground generation in externally defined standard descriptors. We adopt a similar rubric-grounding strategy as a baseline, but study a different problem: simulating and modeling student writing conditioned on target proficiency profiles. Accordingly, our evaluation focuses on profile-level alignment rather than compliance with predefined standards.

\section{SWIM: Student Writing sIMulation}
\label{sec:method}

We study whether language models can simulate student-authored writing aligned with specified proficiency profiles.  We introduce \textbf{SWIM} (Student Writing sIMulation; Figure~\ref{fig:main_fig}), a task that formalizes student writing simulation as {score-guided essay generation} (\S\ref{sec:method-task}). Within this formulation, we examine three families of methods that progressively increase the supervision signal: rubric-grounded prompting (\S\ref{sec:method-prompting}), SFT on score--essay pairs (\S\ref{sec:method-sft}), and GRPO with an AES-derived reward (\S\ref{sec:method-grpo}).

\begin{table}[t]
\label{tab:dataset-overview}
\centering
\small
\setlength{\tabcolsep}{4pt}
\renewcommand{\arraystretch}{1.15}
\scalebox{0.7}{
\begin{tabular}{lcccccc}
\hline
\textbf{Pr} & \textbf{\# Essay} & \textbf{Len} & \textbf{Type} & \textbf{Gr} & \textbf{Traits} & \textbf{Score Range} \\
 &  &  &  &  &  & \textbf{Overall / Trait} \\
\hline
P1 & 1,783 & 350 & Arg.    & 8  & Over, Cont, Org, WC, SF, Conv & 2--12 / 1--6 \\
P2 & 1,800 & 350 & Arg.    & 10 & Over, Cont, Org, WC, SF, Conv & 1--6 / 1--6 \\
P3 & 1,726 & 150 & SD  & 10 & Over, Cont, PA, Lan, Nar      & 0--3 / 0--3 \\
P4 & 1,772 & 150 & SD  & 10 & Over, Cont, PA, Lan, Nar      & 0--3 / 0--3 \\
P5 & 1,805 & 150 & SD  & 8  & Over, Cont, PA, Lan, Nar      & 0--4 / 0--4 \\
P6 & 1,800 & 150 & SD  & 10 & Over, Cont, PA, Lan, Nar      & 0--4 / 0--4 \\
P7 & 1,569 & 300 & Narr.   & 7  & Over, Cont, Org, Conv  & 0--30 / 0--6 \\
P8 &   723 & 650 & Narr.   & 10 & Over, Cont, Org, WC, SF, Conv & 0--60 / 2--12 \\
\hline
\end{tabular}
}
\caption{Prompt-level statistics for the used ASAP/ASAP++~\cite{asap-aes,mathias2018asap++} dataset. Len: average essay length; Gr: grade level; Arg.: argumentative; SD: source-dependent; Narr.: narrative. \textbf{Trait abbreviations:} \textit{Over: Overall, Cont: Content, Org: Organization, WC: Word Choice, SF: Sentence Fluency, Conv: Conventions, PA: Prompt Adherence, Lang: Language, Nar: Narrativity.}}

\label{data}
\end{table}

\subsection{Task Formulation}
\label{sec:method-task}

Let $\mathcal{D} = \{(p_i, x_i, \mathbf{s}_i)\}_{i=1}^{N}$ denote a dataset of writing assessments, where $p_i$ is a writing prompt (including instructions and any source text), $x_i$ is a student-authored essay, and $\mathbf{s}_i = \{s_i^{(t)}\}_{t \in \mathcal{T}_{p_i}}$ is a trait-indexed score vector assigned to $x_i$ by human raters. The set $\mathcal{T}_{p_i}$ contains the scoring dimensions defined for prompt $p_i$ under the task-specific rubric, including the holistic \emph{Overall} score and trait-level dimensions such as \emph{Content} and \emph{Organization} (Table~\ref{data}). Each trait $t$ has a prompt-specific integer score range $[\min_t^{(p)}, \max_t^{(p)}]$.

Given a prompt $p$ and a target proficiency profile $\mathbf{s} = \{s^{(t)}\}_{t \in \mathcal{T}_p}$ taking values in the same space as the rubric scores, the goal of SWIM is to generate an essay $\tilde{x} \sim G_\theta(\cdot \mid p, \mathbf{s})$ that (i) addresses the prompt and (ii) exhibits writing behavior consistent with $\mathbf{s}$. At training time, $\mathbf{s}$ is the observed profile $\mathbf{s}_i$ paired with the gold essay $x_i$; at inference time, $\mathbf{s}$ can be any target profile, including combinations not necessarily present in $\mathcal{D}$.

\paragraph{Input Encoding.}
For all methods, we encode $(p, \mathbf{s})$ as a textual input consisting of (i) a system instruction specifying the role of simulating student writing at a given proficiency level, (ii) a structured trait--score block listing each trait together with its prompt-specific range (\texttt{Trait: score (min--max)}), (iii) the task prompt and any source text, and (iv) an output constraint instructing the model to simulate a student, producing only essay text and not to reference scores or rubric language. This shared interface ensures that the prompting, SFT, and GRPO methods differ only in how $G_\theta$ is obtained, not in the conditioning signal.

\paragraph{Evaluation Interface.}
To measure how closely a generated essay $\tilde{x}$ matches the intended profile, we use a frozen lightweight multi-trait AES model $A_\phi$~\cite{do2024arts} as an external verifier. The verifier maps an essay and prompt to predicted trait scores $\hat{\mathbf{s}} = A_\phi(\tilde{x}, p)$, and we report Quadratic Weighted Kappa (QWK)~\cite{cohen1968weighted} between the target profile $\mathbf{s}$ and the AES-predicted profile $\hat{\mathbf{s}}$, following standard AES practice. We deliberately separate the generator $G_\theta$ from the verifier $A_\phi$. Particularly, $A_\phi$ is fixed across all methods and never updated, so improvements in alignment reflect changes in $G_\theta$'s behavior rather than co-adaptation between generator and evaluator.

\subsection{Rubric-Grounded Prompting}
\label{sec:method-prompting}

We first evaluate whether advanced prompting alone can achieve proficiency-conditioned writing without task-specific parameter updates, using two strategies that progressively incorporate more rubric-derived information into the prompt.

\paragraph{Contrastive Trait Prompting (CTP).}

Following \citet{mannekote2025can}, we construct a contrastive description for each trait that captures how high- and low-proficiency learners differ along that dimension. Descriptions are extracted from the official dataset (in this work, ASAP) grading rubrics and phrased as single high-versus-low contrasts (e.g., for \emph{Content} trait: ``a learner with a higher Content level is more likely to provide clear, focused, thoroughly developed ideas, while a learner with a lower level is more likely to provide undeveloped, minimally focused responses''). To handle cases where the LLM's prior commonsense about \textit{``good writing''} overrides the rubric, the prompt explicitly instructs the model to defer to the contrastive description when the two conflict, following the calibration approach of \citet{mannekote2025can}. The model is then asked to produce an 
essay whose quality aligns with the numerical target scores under these contrastive descriptors.

\paragraph{Score-level Rubric-Lookup Prompting (SRLP).}

CTP exposes only a coarse high/low contrast, which may be insufficient to distinguish intermediate proficiency levels. Building on \citet{imperial2024standardize}, who ground generation in externally defined proficiency descriptors, we adapt the same idea to our grade- and trait-specific rubrics. For each trait $t$ and target score $s^{(t)}$ under prompt $p$, we look up the corresponding rubric descriptor from the official ASAP guidelines 
(e.g., for \emph{Organization} at score 4/6: ``The essay shows satisfactory organization. It contains a basic introduction, body and conclusion.''). The descriptors retrieved for all traits in $\mathcal{T}_p$ are concatenated into the prompt alongside the task instruction. This setting tests whether grounding generation 
in score-specific expert criteria, rather than a binary contrast, yields finer-grained proficiency alignment.

\paragraph{Few-shot Variants.}

We evaluate both CTP and SRLP in five-shot configurations. Five exemplar essays are retrieved from the development set to span the overall-score distribution and are prepended together with their trait-score profiles. This isolates the effect of rubric grounding from that of exposure to authentic student writing examples. Full prompt templates are provided in Appendix~\ref{append:prompt}.

\subsection{Supervised Fine-tuning (SFT)}
\label{sec:method-sft}

The prompting methods rely solely on the base LLM's prior notion of how proficiency is encoded in writing. To test whether direct exposure to real score--essay pairs can induce proficiency-aligned behavior, we train the generator on $\mathcal{D}$ with a standard conditional language-modeling objective:
\begin{equation}
    \mathcal{L}_{\text{SFT}}(\theta) \;=\; 
    -\sum_{i=1}^{N} \log p_\theta(x_i \mid u_i),
\end{equation}
where $u_i$ denotes the input representation of the writing prompt $p_i$ and target score configuration $\mathbf{s}_i$ (\S\ref{sec:method-task}). The model is trained to generate the human-authored essay $x_i$ conditioned on the corresponding proficiency signals provided by human raters.

Unlike prompting strategies, SFT exposes the model to examples of proficiency-conditioned writing, allowing it to directly learn associations between trait-level scores and writing characteristics. Comparing SFT against prompting therefore isolates the value of explicit supervision from naturally occurring score--essay pairs, while subsequent comparison with GRPO reveals whether reward-based optimization can further improve proficiency alignment beyond imitation learning alone.

\subsection{GRPO with Proficiency Alignment Reward (PAR)}
\label{sec:method-grpo}
While SFT learns associations between trait scores and writing characteristics from observed examples, it does not explicitly optimize for proficiency alignment at generation time. Generated essays may remain plausible while still deviating from the intended trait scores. To directly optimize score alignment, we introduce the {Proficiency Alignment Reward} (PAR), a dense per-sample reward that quantifies how closely an essay's AES-predicted trait scores match the target profile. We then 
refine the SFT-initialized policy with GRPO~\cite{shao2024deepseekmath} using PAR as the reward signal.

\paragraph{PAR.}
Let $\pi_\theta$ denote the current policy, and let $\tilde{x} \sim \pi_\theta(\cdot \mid p, \mathbf{s})$ be a sampled essay. We score $\tilde{x}$ with the frozen verifier $A_\phi$ to obtain $\hat{\mathbf{s}} = A_\phi(\tilde{x}, p)$, and define PAR as the mean trait-normalized accuracy across the prompt's trait set:
\begin{equation}
\label{eq:reward}
R(\tilde{x}, \mathbf{s}, p)
=
\frac{1}{|\mathcal{T}_p|}
\sum_{t \in \mathcal{T}_p}
\left(
1-
\frac{
\left|\hat{s}^{(t)}-s^{(t)}\right|
}{
\Delta_t^{(p)}
}
\right).
\end{equation}
where $\Delta_t^{(p)} = \max_t^{(p)} - \min_t^{(p)}$ is the score range of trait $t$ under prompt $p$. PAR is bounded in $[0,1]$, attaining its maximum when all predicted trait scores exactly match the target and decreasing linearly as deviations increase. Each trait contributes equally regardless of its absolute score range.

\paragraph{Policy update.}
For each prompt $p_i$ in a batch, the policy samples a group of $G$ essays $\{\tilde{x}_{i,j}\}_{j=1}^{G}$ conditioned on the same $(p_i, \mathbf{s}_i)$, and computes group-normalized advantages:
\begin{equation}
    A_{i,j} \;=\; 
    \frac{R(\tilde{x}_{i,j}, \mathbf{s}_i, p_i) - \mu_i}
         {\sigma_i + \epsilon},
\end{equation}
with $\mu_i, \sigma_i$ the within-group mean and standard deviation of PAR values. The policy is then updated with the clipped PPO surrogate~\cite{schulman2017proximal}, with no separate value network, following \citet{shao2024deepseekmath}. The reference policy used for the PPO KL term is the SFT checkpoint, snapshotted at the start of GRPO training. Implementation details (group size, learning rate, rollout configuration, and the precision/quantization choices that turn out to matter for training stability) are described in Appendix~\ref{app:result_details}.

\paragraph{Why a dense trait-accuracy reward.}
We deliberately use the smooth trait-normalized accuracy in Eq.~\ref{eq:reward} rather than a discrete ``exact match'' signal or the QWK metric itself. Exact-match rewards are sparse in the multi-trait setting (i.e., most rollouts fail on at least one trait), providing little learning signal. QWK, on the other hand, is defined over a batch of (target, prediction) pairs and is therefore not naturally a per-sample reward~\cite{do2024autoregressive_rl}. PAR provides a dense, per-sample signal that is well-defined for individual rollouts. While it is computed from the same target and predicted trait scores as QWK, it measures their trait-wise closeness for an individual essay rather than agreement over a set of essays.

\definecolor{refgray}{gray}{0.55}
\newcommand{\refcell}[1]{\textcolor{refgray}{#1}}

\begin{table*}[t]
\centering
\setlength{\tabcolsep}{4pt}
\scalebox{0.73}{
\begin{tabular}{cllcccccccccc}
\toprule
& & \textbf{Method} & \textbf{Over.} & \textbf{Cont.} & \textbf{Org.} & \textbf{WC} & \textbf{SF} & \textbf{Conv.} & \textbf{PA} & \textbf{Lang.} & \textbf{Narr.} & \textbf{Avg.} \\
\midrule
\multirow{11}{*}{\rotatebox[origin=c]{90}{\textit{PROMPTING}}}
  & \multirow{4}{*}{Claude Sonnet-4.6}
    & CTP        & 0.488 & 0.540 & 0.400 & 0.480 & 0.446 & 0.310 & 0.632 & 0.644 & 0.652 & 0.510 \\
  & & CTP + FS   & 0.584 & 0.590 & 0.529 & 0.571 & 0.530 & 0.425 & 0.638 & 0.649 & 0.677 & 0.577 \\
  & & SRLP       & 0.487 & 0.526 & 0.380 & 0.564 & 0.530 & 0.321 & 0.636 & 0.760 & 0.692 & 0.544 \\
  & & SRLP + FS  & 0.550 & 0.589 & 0.438 & 0.548 & 0.504 & 0.343 & 0.688 & \textbf{0.769} & 0.733 & 0.574 \\
  \cmidrule{2-13}
  & \multirow{4}{*}{GPT-5.4}
    & CTP        & 0.164 & 0.170 & 0.095 & 0.128 & 0.117 & 0.067 & 0.280 & 0.348 & 0.320 & 0.187 \\
  & & CTP + FS   & 0.296 & 0.263 & 0.173 & 0.162 & 0.173 & 0.113 & 0.391 & 0.429 & 0.437 & 0.277 \\
  & & SRLP       & 0.269 & 0.322 & 0.153 & 0.224 & 0.235 & 0.131 & 0.500 & 0.557 & 0.527 & 0.323 \\
  & & SRLP + FS  & 0.404 & 0.426 & 0.224 & 0.309 & 0.287 & 0.188 & 0.641 & 0.632 & 0.640 & 0.422 \\
  \cmidrule{2-13}
  & \multirow{5}{*}{\refcell{Qwen2.5-7B-Inst$^{\dagger}$}}
    & \refcell{Naive}  & \refcell{0.014} & \refcell{0.001} & \refcell{$-$0.000} & \refcell{0.009} & \refcell{0.005} & \refcell{0.036} & \refcell{$-$0.002} & \refcell{$-$0.002} & \refcell{$-$0.002} & \refcell{0.007} \\
  & & \refcell{CTP}    & \refcell{0.009} & \refcell{$-$0.003} & \refcell{0.022} & \refcell{0.008} & \refcell{0.010} & \refcell{0.028} & \refcell{$-$0.015} & \refcell{$-$0.011} & \refcell{$-$0.003} & \refcell{0.005} \\
& & \refcell{CTP + FS}  & \refcell{0.046} & \refcell{0.019} & \refcell{0.043} & \refcell{0.076} & \refcell{0.049} & \refcell{0.050} & \refcell{0.014} & \refcell{0.014} & \refcell{0.017} & \refcell{0.036} \\
  & & \refcell{SRLP}   & \refcell{0.032} & \refcell{0.025} & \refcell{0.012} & \refcell{0.035} & \refcell{0.024} & \refcell{0.005} & \refcell{0.041} & \refcell{0.030} & \refcell{0.034} & \refcell{0.027} \\
& & \refcell{SRLP + FS} & \refcell{0.177} & \refcell{0.151} & \refcell{0.124} & \refcell{0.178} & \refcell{0.153} & \refcell{0.079} & \refcell{0.174} & \refcell{0.147} & \refcell{0.191} & \refcell{0.153} \\
  
\midrule
\multirow{2}{*}{\rotatebox[origin=c]{90}{\textit{SFT}}}
  & Qwen2.5-7B-Inst & SFT & 0.583 & 0.582 & 0.442 & 0.416 & 0.331 & 0.317 & 0.626 & 0.423 & 0.545 & 0.474 {\scriptsize $\pm$0.023}\\
  & Qwen3-4B        & SFT 
  & 0.437 &	0.498 &	0.356 &	0.421 &	0.389 &	0.327 &	0.532 &	0.383 &	0.508 & 0.428 {\scriptsize $\pm$0.062}\\
\midrule
\multirow{2}{*}{\rotatebox[origin=c]{90}{\textit{RL}}}
  & Qwen2.5-7B-Inst & + GRPO$_{\text{PAR}}$ & \textbf{0.607} & 0.701 & 0.562 & 0.526 & 0.482 & 0.460 & 0.772 & 0.705 & 0.745 & 0.618 {\scriptsize $\pm$0.005} \\
 & Qwen3-4B        & + GRPO$_{\text{PAR}}$ & 0.602 & \textbf{0.726} & \textbf{0.605} & \textbf{0.622} & \textbf{0.606} & \textbf{0.575} & \textbf{0.815} & 0.729 & \textbf{0.789} & \textbf{0.674} {\scriptsize $\pm$0.060}\\
\bottomrule
\end{tabular}
}

\caption{Trait-level QWK between target profiles and AES predictions on ASAP/ASAP++. Prompting uses a single fold; SFT and RL report five-fold means ($\pm$ SD). FS: 5-shot. Bold: best per column. $^{\dagger}$ \textcolor{refgray}{Gray rows} show Qwen2.5-7B-Instruct prompting baselines.}

\label{tab:trait-results}
\end{table*}

\section{Experiments}

\paragraph{Dataset and Experimental Setup.}
We evaluate on ASAP/ASAP++~\cite{mathias2018asap++}, a trait-score-annotated essay corpus derived from the publicly-available Kaggle ASAP dataset,\footnote{\url{https://www.kaggle.com/c/asap-aes}} comprising eight writing prompts that cover argumentative, narrative, and source-based English writing tasks. Each prompt is associated with a prompt-specific set of writing traits and corresponding score ranges. We exclude the Style trait (evaluated only on P7) and the Voice trait (evaluated only on P8), as their single-prompt scope yields insufficient samples for prompt-unified modeling. Following prior AES studies~\cite{taghipour-ng-2016-neural,do2024arts}, we adopt the standard five-fold cross-validation protocol and report average performance across all folds. The same train/validation/test splits are used for both SFT and GRPO experiments. For prompting-based methods, we report results on a single fold due to the substantial computational cost of long-form
essay generation and rubric-conditioned prompting. For each fold, the AES verifier $A_\phi$ is instantiated as the ArTS~\cite{do2024arts} model trained exclusively on that fold's training split, achieving an average prompt-level QWK of $0.722$ on the corresponding test split (full per-prompt verifier scores are reported in Appendix~\ref{app:result_details}). Consequently, the verifier never observes essays from the evaluation set, eliminating evaluation leakage.

\paragraph{Evaluation Protocol.}

To evaluate simulation fidelity, we compare the target trait scores used for conditioning against the scores assigned to generated essays by the AES verifier $A_\phi$. For each test instance $(p_i, x_i, \mathbf{s}_i)$, we condition the generator on the writing prompt $p_i$ and the gold trait-score configuration $\mathbf{s}_i$, and generate an essay $\tilde{x}_i$. The generated essay is subsequently scored by $A_\phi$ to obtain predicted trait scores $\hat{\mathbf{s}}_i$. We report QWK between target and predicted scores at two levels: (i) prompt-level QWK, averaged across traits within each prompt; (ii) trait-level QWK, averaged across prompts containing the corresponding trait. Higher values indicate stronger alignment between intended and generated writing proficiency.

\section{Results}
\label{sec:results}

Tables~\ref{tab:trait-results} and~\ref{tab:prompt-results} report trait-level and prompt-level QWK between target proficiency profiles and AES-predicted profiles. Our analysis investigates how much supervision is required for realistic student writing simulation, progressing from rubric-grounded prompting, to supervised fine-tuning on score--essay pairs, to RL with PAR.

\subsection{Rubric Grounding Helps, but Prompting Alone Is Insufficient}
\label{sec:results-prompting}

Across both Claude Sonnet and GPT~5.4, Score-level SRLP generally outperforms CTP, suggesting that score-specific rubric descriptors provide more effective guidance than coarse high-versus-low trait contrasts. Few-shot demonstrations further yield consistent gains across both prompting strategies, indicating that even a small number of exemplar essays helps models associate target proficiency profiles with student-specific writing patterns. Nevertheless, prompting alone remains insufficient as a general solution for student writing simulation: while Claude Sonnet reaches $0.577$ average trait-level QWK, GPT~5.4 achieves only $0.422$ in its best configuration and open-source Qwen models exhibit near-zero alignment under prompting alone.

\paragraph{Prompting controls what students write, but not how they write.}

\begin{figure}
    \centering
    \includegraphics[width=0.9\linewidth]{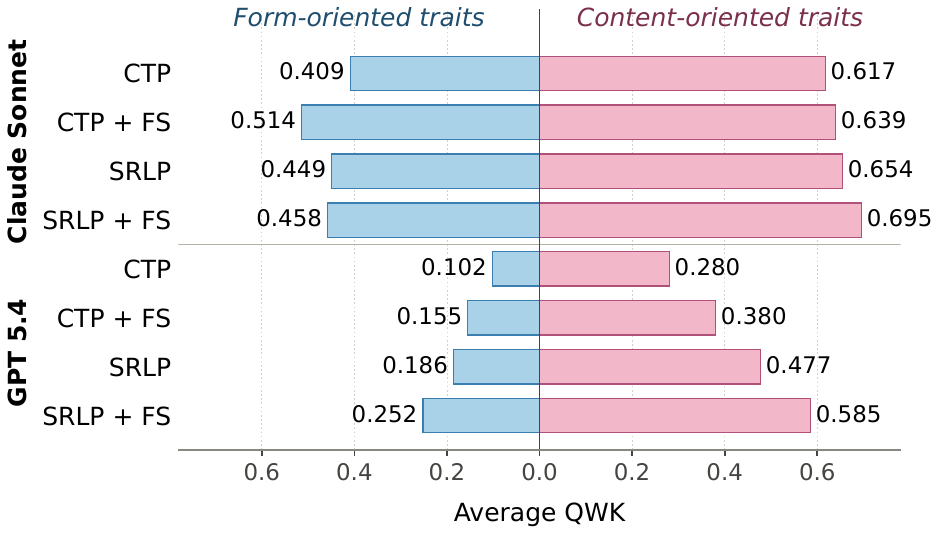}
    \caption{Average QWK on form-oriented traits and content-oriented traits across eight prompting configurations. Content-oriented traits consistently achieve higher alignment than form-oriented traits.}
\label{fig:content-vs-structural}
\end{figure}

Beyond overall averages, a striking pattern emerges when traits are grouped according to whether they primarily assess content or form (Figure~\ref{fig:content-vs-structural}). Across all eight prompting configurations and both proprietary models, content-oriented traits (\emph{Content}, \emph{Prompt Adherence}, \emph{Language}, and \emph{Narrativity}) consistently achieve substantially higher alignment than form-oriented traits (\emph{Organization}, \emph{Word Choice}, \emph{Sentence Fluency}, and \emph{Conventions}). The gap persists across prompting strategies, few-shot demonstrations, and model families, with no exception observed.

Notably, stronger rubric grounding does not eliminate this disparity. Even Claude's best prompting configuration (SRLP + FS) reaches $0.695$ average QWK on content-oriented traits but only $0.458$ on form-oriented traits. Likewise, \emph{Conventions} remains the weakest trait under every prompting setting. The results suggest that prompting can effectively steer models toward different content profiles, but is far less effective at reproducing the lexical and grammatical controls that distinguish weaker and stronger student writers.

We hypothesize that this limitation stems from the alignment objectives of instruction-tuned LLMs, which are explicitly optimized to produce fluent, grammatically correct, and well-structured text. While these models can readily modulate \emph{what} they write (i.e., content), they may struggle to faithfully reproduce \emph{how} lower-proficiency students write (i.e., not merely by introducing errors, but by jointly matching authentic patterns of disfluency, surface errors, and syntactic structure). This inductive bias, consistent with the tendency of aligned LLMs to produce polished, helpful, and well-formed outputs~\citep{ouyang2022training,bai2022constitutional}, is beneficial for most generation tasks, yet may become a fundamental bottleneck when the goal is to simulate lower-proficiency student writing by reproducing its characteristic errors, disfluencies, and structural weaknesses.

\subsection{SFT Enables Proficiency-Conditioned Student Writing}
\label{sec:results-sft}

While prompting relies on rubric descriptions and a few exemplars, SFT exposes the model to large-scale examples of how students at different proficiency levels actually write. To isolate the effect of supervision from differences in backbone and evaluation fold, we compare prompting and SFT using the same Qwen2.5-7B-Instruct backbone on the same fold. SFT achieves an average trait-level QWK of $0.501$, substantially outperforming both CTP+FS ($0.036$) and SRLP+FS ($0.153$). This advantage persists across the full five-fold evaluation: Qwen2.5-7B-Instruct reaches an average QWK of $0.474 \pm 0.023$, while Qwen3-4B achieves $0.428 \pm 0.062$. These results show that substantial proficiency-dependent variation can be learned directly from score--essay pairs, beyond what can be elicited via prompting.

However, proficiency alignment remains uneven across writing dimensions. While models learn to distinguish broader distinctions in writing quality, traits such as \emph{Language}, \emph{Sentence Fluency}, and \emph{Conventions} remain noticeably more challenging than dimensions such as \emph{Content} and \emph{Prompt Adherence}. These findings indicate that exposure to student essays teaches models what different proficiency levels look like, but does not necessarily teach to reproduce the full combination of lexical, grammatical, stylistic, and discourse-level characteristics associated with a target proficiency profile.

\subsection{GRPO with PAR Consistently Improves Proficiency Alignment}
\label{sec:results-grpo}

Refining the policy with GRPO$_{\textsc{PAR}}$ yields consistent improvements over SFT across both backbones, every writing trait, and every essay-prompt (Tables~\ref{tab:trait-results} and~\ref{tab:prompt-results}). Beyond SFT, average trait-wise QWK improves by +0.144 for Qwen2.5-7B-Instruct and +0.246 for Qwen3-4B, while prompt-wise improves by +0.131 and +0.243, respectively. Notably, the SFT-weaker Qwen3-4B benefits most from RL and achieves the strongest overall alignment, suggesting that reward-based optimization can substantially reshape writing behavior beyond what is acquired through supervised imitation. These results indicate that beyond learning what different proficiency levels look like, explicitly optimizing for proficiency alignment yields substantially more faithful student writing simulation.

A noticeable observation is the uniformity of the improvements. Rather than boosting only a few easily controlled dimensions, GRPO improves alignment across all traits and prompts, yielding more balanced proficiency profiles overall. This behavior is consistent with the design of PAR, which rewards agreement with the entire target profile rather than any individual trait. Consequently, the policy learns not only to improve individual traits, but to coordinate multiple writing dimensions simultaneously, producing essays whose overall trait configurations more closely match the requested student's proficiency profile.

\begin{figure}
    \centering
    \includegraphics[width=0.85\linewidth]{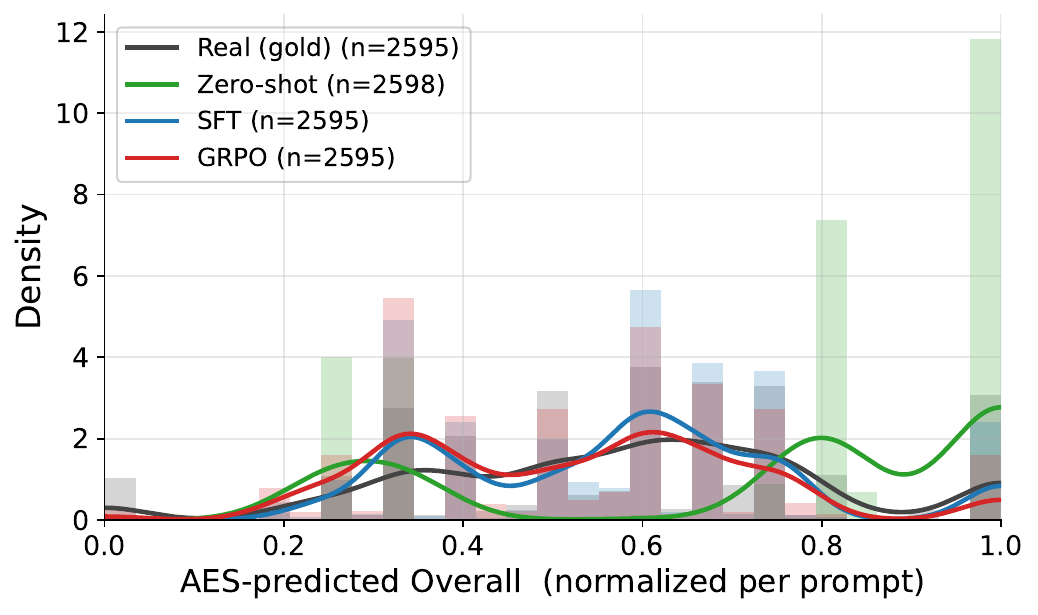}
    \caption{AES-predicted \textit{Overall} score distributions on the test set 
(Qwen2.5-7B-Instruct, fold 1; normalized per prompt to $[0,1]$). Zero-shot prompting 
concentrates near the maximum (mean 0.74, with 25\% of essays receiving 
the cap score), while SFT (0.57) and GRPO (0.53) closely match the human 
distribution (0.58).}
\label{fig:aes-pred-dist}
\label{fig:length}
\end{figure}

\subsection{Do the Gains Transfer to Independent Evaluators?}

As PAR is derived from the same AES verifier used for evaluation, the observed improvements may reflect adaptation to the reward verifier. To assess whether the gains generalize beyond the reward verifier, we conduct an additional analysis on one fold, re-evaluating all generated essays using two independent evaluators that were never involved in policy optimization: (1) an architecturally distinct scorer and (2) an evaluation-only verifier.

We first train a DeBERTa-based multi-trait scorer that predicts rubric scores via classification, providing an architecture distinct from the autoregressive ArTS verifier used during RL. Although trained on the same fold-specific training split, it is never used during RL optimization. In addition, we train an evaluation-only ArTS verifier on the test split, which is unavailable during RL and therefore cannot be adapted to. Because this verifier is fit to the evaluation data, we interpret the relative SFT–GRPO gain rather than its absolute QWK.

Using the DeBERTa scorer, GRPO achieves an average trait-level QWK of 0.598 compared with 0.479 for SFT (+0.119). Using the evaluation-only verifier, the corresponding scores are 0.647 and 0.501 (+0.146), closely matching the original evaluation (0.618 vs. 0.474, +0.144). The consistent improvements across all three independently trained evaluators suggest that the observed gains are not specific to the original reward verifier, but generalize across evaluators the policy never interacted with during training.
\section{Discussion}\label{sec6}

\paragraph{Why is writing simulation difficult?}

Despite the strong gains from SFT and GRPO, writing simulation remains challenging in certain settings. The clearest example is Prompt~8 (Table~\ref{tab:prompt-results}), where prompting-based methods outperform both SFT and GRPO. While Qwen3-4B + GRPO achieves the strongest average prompt-level alignment overall, its performance on Prompt~8 remains substantially below Claude Sonnet with SRLP.

One possible explanation is the distinctive data distribution of this prompt. Prompt~8 contains only 723 training essays (less than half the size of most other prompts) while requiring substantially longer responses (average length $\approx650$). This combination forces the model to learn proficiency-dependent writing behavior from limited supervision while maintaining the target proficiency profile over long-form generations. The results suggest that data scarcity and generation length remain important bottlenecks for student writing simulation, even under explicit profile-level optimization.

\begin{figure}
    \centering
    \includegraphics[width=0.85\linewidth]{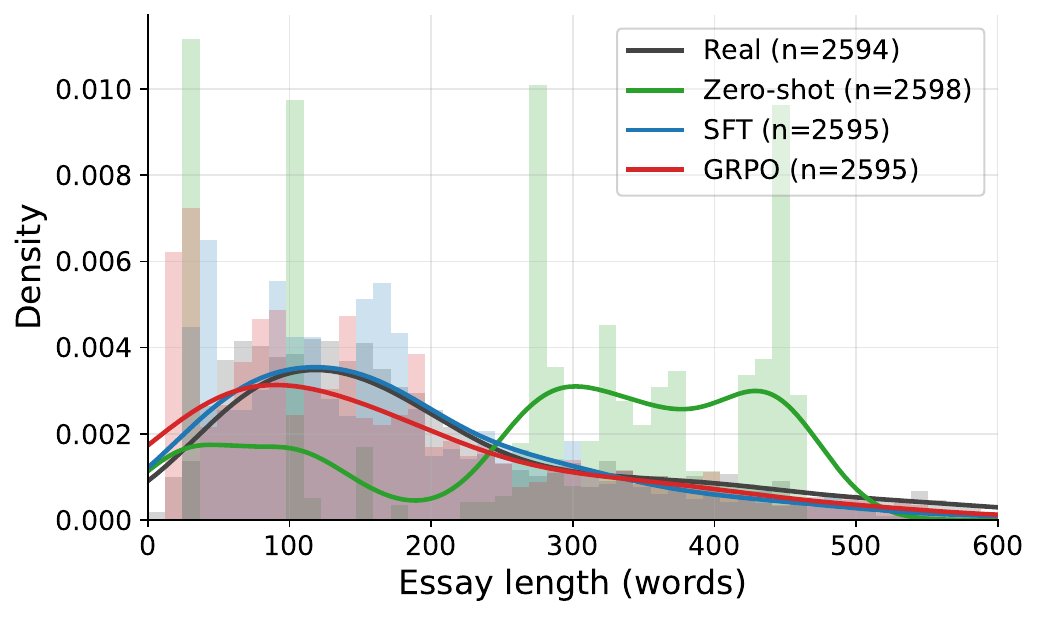}
    \caption{Essay-length distributions on the test set (Qwen2.5-7B-Instruct, fold 1). 
    }
\label{fig:length}
\end{figure}

\paragraph{Prompting collapses proficiency variation.}

Figure~\ref{fig:aes-pred-dist} compares the distribution of AES-predicted overall scores against the real student population. Zero-shot prompting exhibits a strong bias toward high-scoring essays, with a mean normalized score of 0.74 and 25\% of generations receiving the maximum score. In contrast, SFT (0.57) and GRPO (0.53) closely match the human distribution (0.58). These results suggest that prompting alone tends to simulate an ``idealized'' (i.e., high-performing) student population, whereas proficiency-aware supervision enables models to reproduce a broader spectrum of writing ability.

\paragraph{Student simulators learn realistic writing length.}

A similar pattern emerges when examining essay length distributions (Figure~\ref{fig:length}). Zero-shot prompting produces substantially longer essays than real students (median 304 vs.\ 167 words), reflecting the well-known tendency of instruction-tuned LLMs toward verbose and polished responses~\cite{singhal2023long,saito2023verbosity}. In contrast, SFT (158) and GRPO (141) closely recover the human distribution despite receiving no explicit length supervision. This suggests that proficiency-aware training captures broader characteristics of student writing beyond trait-level score alignment alone.

\begin{table}[t]
\centering
\small
\setlength{\tabcolsep}{4pt}
\begin{tabular}{lrrrr}
\toprule
\textbf{Metric}
& \textbf{Real}
& \textbf{Claude-CTP}
& \textbf{SFT}
& \textbf{GRPO} \\
\midrule

Grammar /100w
& \cellcolor{gray!15}1.04
& \cellcolor{blue!25}\textbf{1.15}
& \cellcolor{blue!7}0.16
& \cellcolor{blue!7}0.00 \\

Spelling /100w
& \cellcolor{gray!15}4.22
& \cellcolor{blue!15}\textbf{6.31}
& \cellcolor{blue!7}1.01
& \cellcolor{blue!7}0.71 \\

MATTR ($w=50$)
& \cellcolor{gray!15}0.76
& \cellcolor{blue!35}\textbf{0.75}
& \cellcolor{blue!25}0.68
& \cellcolor{blue!35}0.73 \\

Casing /100w
& \cellcolor{gray!15}0.65
& 12.57
& \cellcolor{blue!7}\textbf{0.00}
& \cellcolor{blue!7}\textbf{0.00} \\

Clauses / sentence
& \cellcolor{gray!15}3.15
& \cellcolor{blue!7}1.52
& \cellcolor{blue!25}3.51
& \cellcolor{blue!35}\textbf{3.45} \\

Dependency depth
& \cellcolor{gray!15}5.34
& \cellcolor{blue!7}2.57
& \cellcolor{blue!7}8.09
& \cellcolor{blue!35}\textbf{5.48} \\

\bottomrule
\end{tabular}
\caption{Linguistic characteristics of real and generated essays under low-proficiency targets. Error counts are per 100 words. Darker blue denotes smaller relative deviation from real essays within each metric; bold marks the closest generated value. Details in Appendix~\ref{app:linguistic-metrics}}

\label{tab:linguistic_analysis}
\end{table}

\paragraph{LLMs struggle to reproduce low-proficiency linguistic forms.}

Our results in §~\ref{sec:results-prompting} suggest that models can produce high-proficiency writing more easily than authentic low-proficiency writing, which requires reproducing errors and weaker linguistic structures. To characterize how this limitation manifests across generation methods, we compare 247 low-target essays per source (\textit{Overall} $\leq 0.33$ of the prompt-normalized range) with a prompt-matched high-target cohort of 245 essays per source (\textit{Overall} $\geq 0.67$), measuring grammatical, spelling, and casing errors, lexical diversity, and syntactic complexity. At low targets, Claude CTP approximates the aggregate grammatical-error rate of real essays, but primarily simulates low proficiency through excessive casing and spelling errors while producing substantially simpler syntax (Table~\ref{tab:linguistic_analysis}). In contrast, SFT and GRPO better recover lexical and structural properties, particularly under GRPO, but generate unrealistically few grammatical and spelling errors. The discrepancy becomes substantially smaller at high proficiency; for example, SFT's grammatical-error gap decreases from 84\% below the real rate at low targets to 5\% below it at high targets. These results reveal distinct failure modes: prompting tends to simulate low proficiency through superficial corruption, whereas training recovers syntactic structure but retains overly polished surface realization. Thus, the central challenge is not simply generating less fluent text, but reproducing authentic and proficiency-calibrated patterns of linguistic form.

\paragraph{No evidence of direct memorization or generation collapse.}

We further examine whether the strong alignment of SFT and GRPO could arise from training-set memorization or reduced generation diversity. For non-source-based prompts, neither method produces any near-duplicate of a training essay (0.00\% with ROUGE-L $>0.7$), and their 8-gram overlap is negligible and comparable to that of held-out real essays ($\leq 0.12\%$ vs.\ $0.09\%$). For source-based prompts, overlap is naturally higher because both real and generated essays draw on shared passages; nevertheless, maximum pairwise overlap remains low (6--7\%), and manual inspection identifies no generated essay matching an individual training essay. We also group generations by identical prompt and full trait profile, yielding 132 groups with an average of 11.6 essays each. No essay is repeated within any group. These analyses support the interpretation of the alignment gains as learned proficiency-conditioned behavior rather than memorization or output collapse.

\section{Conclusion}

We introduced SWIM, a task that formulates Student Writing sIMulation as proficiency-conditioned essay generation. Our comparison reveals a clear trend: rubric-grounded prompting offers limited proficiency control even for strong proprietary models, SFT learns substantially stronger alignment from real score--essay pairs, and GRPO with the proposed PAR improves further across every writing trait and essay prompt. Yet profile alignment does not guarantee behavioral realism: across all methods, models reproduce high-proficiency writing more readily than authentic low-proficiency writing, exhibiting distinct failure modes. This may reflect instruction-tuned LLMs' fluency prior, favoring fluent and well-formed text. The next challenge is thus not tighter score alignment alone, but faithfully reproducing the proficiency-calibrated ways students actually write. SWIM takes a step toward student writing simulation that could support the training and evaluation of educational AI across diverse learner profiles.

\section*{Limitations}

This work represents an initial study of student writing simulation and has several limitations. First, our evaluation relies on a state-of-the-art AES model as a proxy for simulation fidelity. Although AES provides scalable and fine-grained trait-level assessment, it cannot fully replace human judgment of whether generated essays authentically resemble student writing. Future work should incorporate expert evaluation and broader measures of behavioral realism. 
Second, our experiments are conducted on ASAP/ASAP++, which covers a limited set of writing prompts and proficiency distributions. While this benchmark provides standardized trait-level annotations, it remains unclear whether the observed findings generalize to other educational contexts, age groups, languages, or assessment frameworks. 
Finally, we model student writing through static proficiency profiles. Real learners may exhibit richer variation, including evolving knowledge, misconceptions, motivation, and writing strategies over time. Extending writing simulation to dynamic learner states and longitudinal writing behavior remains an important direction for future research.

\section*{Ethical Statement}

This work studies student writing simulation using the publicly available ASAP/ASAP++ datasets. The essays are used solely for research purposes and do not contain personal or sensitive student information. AI assistance was employed for language editing and proofreading.

\section*{Acknowledgments}
This research was primarily supported by the ETH AI Center through an ETH AI Center postdoctoral fellowship to Heejin Do. It was also supported by the Swiss National Science Foundation (SNSF) under grant number 10009282 and by a Swiss AI large grant.

\bibliography{custom}

@inproceedings{wang-etal-2022-use,
    title = "On the Use of Bert for Automated Essay Scoring: Joint Learning of Multi-Scale Essay Representation",
    author = "Wang, Yongjie  and
      Wang, Chuang  and
      Li, Ruobing  and
      Lin, Hui",
    editor = "Carpuat, Marine  and
      de Marneffe, Marie-Catherine  and
      Meza Ruiz, Ivan Vladimir",
    booktitle = "Proceedings of the 2022 Conference of the North American Chapter of the Association for Computational Linguistics: Human Language Technologies",
    month = jul,
    year = "2022",
    address = "Seattle, United States",
    publisher = "Association for Computational Linguistics",
    url = "https://aclanthology.org/2022.naacl-main.249/",
    doi = "10.18653/v1/2022.naacl-main.249",
    pages = "3416--3425"
}

@inproceedings{lee-etal-2024-unleashing,
    title = "Unleashing Large Language Models' Proficiency in Zero-shot Essay Scoring",
    author = "Lee, Sanwoo  and
      Cai, Yida  and
      Meng, Desong  and
      Wang, Ziyang  and
      Wu, Yunfang",
    editor = "Al-Onaizan, Yaser  and
      Bansal, Mohit  and
      Chen, Yun-Nung",
    booktitle = "Findings of the Association for Computational Linguistics: EMNLP 2024",
    month = nov,
    year = "2024",
    address = "Miami, Florida, USA",
    publisher = "Association for Computational Linguistics",
    url = "https://aclanthology.org/2024.findings-emnlp.10/",
    doi = "10.18653/v1/2024.findings-emnlp.10",
    pages = "181--198"
}

@inproceedings{chu-etal-2025-rationale,
    title = "Rationale Behind Essay Scores: Enhancing {S}-{LLM}{'}s Multi-Trait Essay Scoring with Rationale Generated by {LLM}s",
    author = "Chu, SeongYeub  and
      Kim, Jong Woo  and
      Wong, Bryan  and
      Yi, Mun Yong",
    editor = "Chiruzzo, Luis  and
      Ritter, Alan  and
      Wang, Lu",
    booktitle = "Findings of the Association for Computational Linguistics: NAACL 2025",
    month = apr,
    year = "2025",
    address = "Albuquerque, New Mexico",
    publisher = "Association for Computational Linguistics",
    url = "https://aclanthology.org/2025.findings-naacl.322/",
    doi = "10.18653/v1/2025.findings-naacl.322",
    pages = "5811--5829",
    ISBN = "979-8-89176-195-7"
}

@article{DO2026132119,
title = {Teach-to-reason with scoring: Self-explainable rationale-driven multi-trait essay scoring},
journal = {Expert Systems with Applications},
volume = {319},
pages = {132119},
year = {2026},
issn = {0957-4174},
doi = {https://doi.org/10.1016/j.eswa.2026.132119},
url = {https://www.sciencedirect.com/science/article/pii/S0957417426010328},
author = {Heejin Do and Sangwon Ryu and Gary Geunbae Lee}
}

@book{gregg2016cognitive,
  title={Cognitive processes in writing},
  author={Gregg, Lee W and Steinberg, Erwin R},
  year={2016},
  publisher={Routledge}
}

@article{saito2023verbosity,
  title={Verbosity bias in preference labeling by large language models},
  author={Saito, Keita and Wachi, Akifumi and Wataoka, Koki and Akimoto, Youhei},
  journal={arXiv preprint arXiv:2310.10076},
  year={2023}
}

@book{bereiter2013psychology,
  title={The psychology of written composition},
  author={Bereiter, Carl and Scardamalia, Marlene},
  year={2013},
  publisher={Routledge}
}

@article{singhal2023long,
  title={A long way to go: Investigating length correlations in rlhf},
  author={Singhal, Prasann and Goyal, Tanya and Xu, Jiacheng and Durrett, Greg},
  journal={arXiv preprint arXiv:2310.03716},
  year={2023}
}

@inproceedings{kwon2023efficient,
  title={Efficient memory management for large language model serving with pagedattention},
  author={Kwon, Woosuk and Li, Zhuohan and Zhuang, Siyuan and Sheng, Ying and Zheng, Lianmin and Yu, Cody Hao and Gonzalez, Joseph and Zhang, Hao and Stoica, Ion},
  booktitle={Proceedings of the 29th symposium on operating systems principles},
  pages={611--626},
  year={2023}
}

@misc{vonwerra2022trl,
  author = {Leandro von Werra and Younes Belkada and Lewis Tunstall and Edward Beeching and Tristan Thrush and Nathan Lambert and Shengyi Huang and Kashif Rasul and Quentin Gallouédec},
  title = {TRL: Transformer Reinforcement Learning},
  year = {2020},
  publisher = {GitHub},
  journal = {GitHub repository},
  howpublished = {\url{https://github.com/huggingface/trl}}
}

@article{dettmers2023qlora,
  title={Qlora: Efficient finetuning of quantized llms},
  author={Dettmers, Tim and Pagnoni, Artidoro and Holtzman, Ari and Zettlemoyer, Luke},
  journal={Advances in neural information processing systems},
  volume={36},
  pages={10088--10115},
  year={2023}
}

@article{honnibal2020spacy,
  title={spaCy: Industrial-strength natural language processing in Python},
  author={Honnibal, Matthew and Montani, Ines and Van Landeghem, Sofie and Boyd, Adriane and others},
  year={2020},
  publisher={Zenodo, Honolulu, HI, USA}
}

@article{ouyang2022training,
  title={Training language models to follow instructions with human feedback},
  author={Ouyang, Long and Wu, Jeffrey and Jiang, Xu and Almeida, Diogo and Wainwright, Carroll and Mishkin, Pamela and Zhang, Chong and Agarwal, Sandhini and Slama, Katarina and Ray, Alex and others},
  journal={Advances in neural information processing systems},
  volume={35},
  pages={27730--27744},
  year={2022}
}

@article{lu2010automatic,
  title={Automatic analysis of syntactic complexity in second language writing},
  author={Lu, Xiaofei},
  journal={International journal of corpus linguistics},
  volume={15},
  number={4},
  pages={474--496},
  year={2010},
  publisher={John Benjamins}
}

@article{covington2010cutting,
  title={Cutting the Gordian knot: The moving-average type--token ratio (MATTR)},
  author={Covington, Michael A and McFall, Joe D},
  journal={Journal of quantitative linguistics},
  volume={17},
  number={2},
  pages={94--100},
  year={2010},
  publisher={Taylor \& Francis}
}

@article{bai2022constitutional,
  title={Constitutional ai: Harmlessness from ai feedback},
  author={Bai, Yuntao and Kadavath, Saurav and Kundu, Sandipan and Askell, Amanda and Kernion, Jackson and Jones, Andy and Chen, Anna and Goldie, Anna and Mirhoseini, Azalia and McKinnon, Cameron and others},
  journal={arXiv preprint arXiv:2212.08073},
  year={2022}
}

@inproceedings{mannekote2025can,
  title={Can llms reliably simulate human learner actions? a simulation authoring framework for open-ended learning environments},
  author={Mannekote, Amogh and Davies, Adam and Kang, Jina and Boyer, Kristy Elizabeth},
  booktitle={Proceedings of the AAAI Conference on Artificial Intelligence},
  volume={39},
  number={28},
  pages={29044--29052},
  year={2025}
}

@article{xu2023simulate,
  title   = {Leveraging Generative Artificial Intelligence to 
             Simulate Student Learning Behavior},
  author  = {Xu, Songlin and Zhang, Xinyu},
  journal = {arXiv preprint arXiv:2310.19206},
  year    = {2023}
}

@inproceedings{lu2024generative,
  title     = {Generative Students: Using {LLM}-Simulated Student 
               Profiles to Support Question Item Evaluation},
  author    = {Lu, Xinyi and Wang, Xu},
  booktitle = {Proceedings of the Eleventh ACM Conference on 
               Learning @ Scale (L@S '24)},
  year      = {2024}
}

@inproceedings{nair2024closing,
  title     = {Closing the Loop: Learning to Generate Writing 
               Feedback via Language Model Simulated Student 
               Revisions},
  author    = {Nair, Inderjeet Jayakumar and Tan, Jiaye and Su, 
               Xiaotian and Gere, Anne and Wang, Xu and Wang, Lu},
  booktitle = {Proceedings of the 2024 Conference on Empirical 
               Methods in Natural Language Processing (EMNLP)},
  pages     = {16636--16657},
  year      = {2024}
}

@inproceedings{benedetto2024llmexam,
  title     = {Using {LLMs} to Simulate Students' Responses to 
               Exam Questions},
  author    = {Benedetto, Luca and Aradelli, Giovanni and Donvito, 
               Antonia and Lucchetti, Alberto and Cappelli, Andrea 
               and Buttery, Paula},
  booktitle = {Findings of the Association for Computational 
               Linguistics: EMNLP 2024},
  year      = {2024}
}

@incollection{kass1989student,
  title={Student modeling in intelligent tutoring systems—implications for user modeling},
  author={Kass, Robert},
  booktitle={User models in dialog systems},
  pages={386--410},
  year={1989},
  publisher={Springer}
}

@article{pavlik2013review,
  title   = {A Review of Student Models Used in Intelligent 
             Tutoring Systems},
  author  = {Pavlik, Philip I. and Brawner, Keith and Olney, 
             Andrew and Mitrovic, Antonija},
  journal = {Design Recommendations for Intelligent Tutoring 
             Systems},
  volume  = {1},
  pages   = {39--68},
  year    = {2013}
}

@article{uto2021review,
  title   = {A Review of Deep-Neural Automated Essay Scoring Models},
  author  = {Uto, Masaki},
  journal = {Behaviormetrika},
  volume  = {48},
  number  = {2},
  pages   = {459--484},
  year    = {2021}
}

@inproceedings{dong2017attention,
  title     = {Attention-Based Recurrent Convolutional Neural 
               Network for Automatic Essay Scoring},
  author    = {Dong, Fei and Zhang, Yue and Yang, Jie},
  booktitle = {Proceedings of the 21st Conference on Computational 
               Natural Language Learning (CoNLL 2017)},
  pages     = {153--162},
  year      = {2017}
}

@inproceedings{taghipour-ng-2016-neural,
    title = "A Neural Approach to Automated Essay Scoring",
    author = "Taghipour, Kaveh  and
      Ng, Hwee Tou",
    editor = "Su, Jian  and
      Duh, Kevin  and
      Carreras, Xavier",
    booktitle = "Proceedings of the 2016 Conference on Empirical Methods in Natural Language Processing",
    month = nov,
    year = "2016",
    address = "Austin, Texas",
    publisher = "Association for Computational Linguistics",
    url = "https://aclanthology.org/D16-1193/",
    doi = "10.18653/v1/D16-1193",
    pages = "1882--1891"
}

@inproceedings{li2024aes,
  title     = {Automated Essay Scoring: Recent Successes and Future 
               Directions},
  author    = {Li, Shengjie and Ng, Vincent},
  booktitle = {Proceedings of the 33rd International Joint 
               Conference on Artificial Intelligence (IJCAI)},
  year      = {2024}
}

@inproceedings{do2024arts,
  title     = {Autoregressive Score Generation for Multi-Trait 
               Essay Scoring},
  author    = {Do, Heejin and Kim, Yunsu and Lee, Gary},
  booktitle = {Findings of the Association for Computational 
               Linguistics: EACL 2024},
  pages     = {1659--1666},
  year      = {2024}
}

@inproceedings{do2024autoregressive_rl,
  title={Autoregressive multi-trait essay scoring via reinforcement learning with scoring-aware multiple rewards},
  author={Do, Heejin and Ryu, Sangwon and Lee, Gary},
  booktitle={Proceedings of the 2024 Conference on Empirical Methods in Natural Language Processing},
  pages={16427--16438},
  year={2024}
}

@article{shao2024deepseekmath,
  title={Deepseekmath: Pushing the limits of mathematical reasoning in open language models},
  author={Shao, Zhihong and Wang, Peiyi and Zhu, Qihao and Xu, Runxin and Song, Junxiao and Bi, Xiao and Zhang, Haowei and Zhang, Mingchuan and Li, YK and Wu, Yang and others},
  journal={arXiv preprint arXiv:2402.03300},
  year={2024}
}

@article{schulman2017proximal,
  title={Proximal policy optimization algorithms},
  author={Schulman, John and Wolski, Filip and Dhariwal, Prafulla and Radford, Alec and Klimov, Oleg},
  journal={arXiv preprint arXiv:1707.06347},
  year={2017}
}

@article{cohen1968weighted,
  title={Weighted kappa: nominal scale agreement provision for scaled disagreement or partial credit.},
  author={Cohen, Jacob},
  journal={Psychological bulletin},
  volume={70},
  number={4},
  pages={213},
  year={1968},
  publisher={American Psychological Association}
}

@inproceedings{mathias2018asap++,
  title={ASAP++: Enriching the ASAP automated essay grading dataset with essay attribute scores},
  author={Mathias, Sandeep and Bhattacharyya, Pushpak},
  booktitle={Proceedings of the eleventh international conference on language resources and evaluation (LREC 2018)},
  year={2018}
}

@article{attali2006erater,
  title={Automated essay scoring with e-rater{\textregistered} V. 2},
  author={Attali, Yigal and Burstein, Jill},
  journal={The Journal of Technology, Learning and Assessment},
  volume={4},
  number={3},
  year={2006}
}

@article{ramineni2012evaluation,
  title={Evaluation of the e-rater{\textregistered} Scoring Engine for the TOEFL{\textregistered} Independent and Integrated Prompts},
  author={Ramineni, Chaitanya and Trapani, Catherine S and Williamson, David M and Davey, Tim and Bridgeman, Brent},
  journal={ETS Research Report Series},
  volume={2012},
  number={1},
  pages={i--51},
  year={2012},
  publisher={Wiley Online Library}
}

@article{liang2024ctgsurvey,
  title={Controllable text generation for large language models: A survey},
  author={Liang, Xun and Wang, Hanyu and Wang, Yezhaohui and Song, Shichao and Yang, Jiawei and Niu, Simin and Hu, Jie and Liu, Dan and Yao, Shunyu and Xiong, Feiyu and others},
  journal={arXiv preprint arXiv:2408.12599},
  year={2024}
}

@article{zhang2023macsum,
  title={Macsum: Controllable summarization with mixed attributes},
  author={Zhang, Yusen and Liu, Yang and Yang, Ziyi and Fang, Yuwei and Chen, Yulong and Radev, Dragomir and Zhu, Chenguang and Zeng, Michael and Zhang, Rui},
  journal={Transactions of the Association for Computational Linguistics},
  volume={11},
  pages={787--803},
  year={2023},
  publisher={MIT Press One Broadway, 12th Floor, Cambridge, Massachusetts 02142, USA~…}
}

@inproceedings{urlana2024controllable,
  title     = {Controllable Text Summarization: Unraveling 
               Challenges, Approaches, and Prospects -- A Survey},
  author    = {Urlana, Ashok and Mishra, Pruthwik and Roy, 
               Tathagato and Mishra, Rahul},
  booktitle = {Findings of the Association for Computational 
               Linguistics: ACL 2024},
  pages     = {1603--1623},
  year      = {2024}
}

@inproceedings{liu2024benchmarking,
  title     = {Benchmarking Generation and Evaluation Capabilities 
               of Large Language Models for Instruction 
               Controllable Summarization},
  author    = {Liu, Yixin and Fabbri, Alexander Richard and Chen, 
               Jiawen and Zhao, Yilun and Han, Simeng and Joty, 
               Shafiq and Liu, Pengfei and Radev, Dragomir and Wu, 
               Chien-Sheng and Cohan, Arman},
  booktitle = {Findings of the Association for Computational 
               Linguistics: NAACL 2024},
  pages     = {4481--4501},
  year      = {2024}
}

@inproceedings{ryu2024iterative,
  title={Exploring iterative controllable summarization with large language models},
  author={Ryu, Sangwon and Do, Heejin and Kim, Daehui and Yu, Hwanjo and Kim, Dongwoo and Kim, Yunsu and Lee, Gary and Ok, Jungseul},
  booktitle={Findings of the Association for Computational Linguistics: EACL 2026},
  pages={512--528},
  year={2026}
}

@inproceedings{li2024tole,
  title={Reinforcement learning with token-level feedback for controllable text generation},
  author={Li, Wendi and Wei, Wei and Xu, Kaihe and Xie, Wenfeng and Chen, Dangyang and Cheng, Yu},
  booktitle={Findings of the Association for Computational Linguistics: NAACL 2024},
  pages={1704--1719},
  year={2024}
}

@inproceedings{imperial2024standardize,
  title={Standardize: Aligning language models with expert-defined standards for content generation},
  author={Imperial, Joseph Marvin and Forey, Gail and Madabushi, Harish Tayyar},
  booktitle={Proceedings of the 2024 Conference on Empirical Methods in Natural Language Processing},
  pages={1573--1594},
  year={2024}
}

@misc{asap-aes,
    author = {Ben Hamner and Jaison Morgan and lynnvandev and Mark Shermis and Tom Vander Ark},
    title = {The Hewlett Foundation: Automated Essay Scoring},
    year = {2012},
    howpublished = {\url{https://kaggle.com/competitions/asap-aes}},
    note = {Kaggle}
}

@inproceedings{scarlatos-etal-2025-smart,
    title = "{SMART}: Simulated Students Aligned with Item Response Theory for Question Difficulty Prediction",
    author = "Scarlatos, Alexander  and
      Fernandez, Nigel  and
      Ormerod, Christopher  and
      Lottridge, Susan  and
      Lan, Andrew",
    editor = "Christodoulopoulos, Christos  and
      Chakraborty, Tanmoy  and
      Rose, Carolyn  and
      Peng, Violet",
    booktitle = "Proceedings of the 2025 Conference on Empirical Methods in Natural Language Processing",
    month = nov,
    year = "2025",
    address = "Suzhou, China",
    publisher = "Association for Computational Linguistics",
    url = "https://aclanthology.org/2025.emnlp-main.1274/",
    doi = "10.18653/v1/2025.emnlp-main.1274",
    pages = "25071--25094",
    ISBN = "979-8-89176-332-6"
}

\newpage
\appendix

\section{Implementation Details}
\label{append:detail}

For SFT, we fine-tune Qwen2.5-7B-Instruct and Qwen3-4B using QLoRA~\citep{dettmers2023qlora} with NF4 4-bit quantization, bf16 computation, and LoRA adapters (rank 16, $\alpha=32$, dropout 0.05) applied to all attention and MLP projections. We use AdamW with a peak learning rate of $10^{-4}$, cosine decay with 10\% warmup, effective batch size 8, maximum sequence length 2500, and train for 5 epochs. The AES verifier is the fold-specific ArTS checkpoint. For RL, we initialize from the SFT checkpoint and optimize with TRL's \texttt{GRPOTrainer}~\citep{vonwerra2022trl} using vLLM colocate rollout~\citep{kwon2023efficient}. We use 4 generations per prompt, learning rate $10^{-5}$, bf16, effective batch size 16, cosine decay with 3\% warmup, and train for one epoch. Maximum prompt and completion lengths are 2500 and 1024 tokens, respectively, and the SFT checkpoint serves as the KL reference policy. At evaluation time, essays are scored using the same fold-specific ArTS verifier. All experiments are conducted on a single NVIDIA H100 (96GB); one SFT epoch requires approximately 1 hour, while one GRPO epoch requires approximately 8 hours.

\section{Additional Results}
\label{app:result_details}
\paragraph{Prompt-wise Results.}

Table~\ref{tab:prompt-results} reports proficiency-alignment performance for each ASAP/ASAP++ essay prompt. The prompt-level trends are consistent with the trait-level analysis in the main text: prompting methods provide limited control, SFT substantially improves alignment over naive prompting on the same backbone, and GRPO$_{\textsc{PAR}}$ achieves the strongest performance across most prompts. We additionally observe greater variation across prompts than across traits, suggesting that prompt characteristics such as genre, length, and training set size influence simulation difficulty; most notably, P8 (long narratives, smallest training set) consistently underperforms across all methods.

\begin{table}[t]
\centering
\small
\setlength{\tabcolsep}{3pt}
\begin{minipage}{0.49\columnwidth}
\centering
\scalebox{0.82}{
\begin{tabular}{lc}
\toprule
\textbf{Trait} & \textbf{QWK} \\
\midrule
Overall          & 0.756 $\pm$ 0.008 \\
Content          & 0.734 $\pm$ 0.011 \\
Organization     & 0.677 $\pm$ 0.022 \\
Word Choice      & 0.685 $\pm$ 0.018 \\
Sentence Fluency & 0.677 $\pm$ 0.016 \\
Conventions      & 0.685 $\pm$ 0.011 \\
Prompt Adherence & 0.757 $\pm$ 0.018 \\
Language         & 0.691 $\pm$ 0.018 \\
Narrativity      & 0.731 $\pm$ 0.017 \\
\midrule
\textbf{Avg.}    & \textbf{0.710 $\pm$ 0.008} \\
\bottomrule
\end{tabular}
}
\end{minipage}
\hfill
\begin{minipage}{0.49\columnwidth}
\centering
\scalebox{0.82}{
\begin{tabular}{lc}
\toprule
\textbf{Prompt} & \textbf{QWK} \\
\midrule
P1 & 0.704 $\pm$ 0.028 \\
P2 & 0.696 $\pm$ 0.023 \\
P3 & 0.710 $\pm$ 0.034 \\
P4 & 0.776 $\pm$ 0.008 \\
P5 & 0.724 $\pm$ 0.024 \\
P6 & 0.772 $\pm$ 0.012 \\
P7 & 0.762 $\pm$ 0.029 \\
P8 & 0.635 $\pm$ 0.039 \\
\midrule
\textbf{Avg.}    & \textbf{0.722 $\pm$ 0.008} \\
\bottomrule
\end{tabular}
}
\end{minipage}
\caption{Performance of the AES verifier ($A_\phi$, 
ArTS~\citep{do2024arts}) implemented and used in our experiments, averaged across 
five cross-validation folds. Left: trait-level QWK. Right: 
prompt-level QWK.}
\label{tab:scorer-performance}
\end{table}

\paragraph{Qualitative Examples.}
Table~\ref{tab:qual-grpo} shows three essays generated by GRPO-trained Qwen2.5-7B-Instruct for the same writing prompt (ASAP Prompt~5, ``mood in \emph{Narciso Rodriguez}'') but under different target Overall scores (1/4, 2/4, 4/4). Several qualitative differences emerge that align with the requested proficiency levels. At the lowest target (1/4), the generated essay consists of two short, generic sentences without any engagement with the source text. At the intermediate level (2/4), the essay grows substantially in length and develops a clearer claim about the author's parents, but its supporting details remain repetitive and largely paraphrased. At the highest target (4/4), the essay explicitly names the memoir and its author, draws on a direct quotation from the source text, and develops the argument through multiple supporting examples. Both length and structural sophistication scale monotonically with the requested proficiency level, and the AES verifier recovers the target Overall score in each case ($\hat{s} = 1.0, 2.0, 4.0$). These examples illustrate that GRPO with PAR produces qualitatively distinct writing behaviors across
proficiency levels rather than merely shifting surface-level statistics.

\paragraph{AES Verifier Performance.}

Table~\ref{tab:scorer-performance} reports the trait-level and prompt-level QWK of the ArTS~\citep{do2024arts} AES verifier $A_\phi$ used throughout our experiments, averaged across five cross-validation folds. The verifier achieves average QWK of $0.710$ at the trait level and $0.722$ at the prompt level, indicating substantial agreement with human raters.

\begin{table*}[t]
\centering
\setlength{\tabcolsep}{6pt}
\scalebox{0.73}{
\begin{tabular}{cllccccccccc}
\toprule
& & \textbf{Method} & \textbf{P1} & \textbf{P2} & \textbf{P3} & \textbf{P4} & \textbf{P5} & \textbf{P6} & \textbf{P7} & \textbf{P8} & \textbf{Avg.} \\
\midrule
\multirow{11}{*}{\rotatebox[origin=c]{90}{\textit{PROMPTING}}}
  & \multirow{4}{*}{Claude Sonnet-4.6}
    & CTP        & 0.618 & 0.540 & 0.664 & 0.649 & 0.505 & 0.650 & 0.331 & 0.157 & 0.514 \\
  & & CTP + FS   & 0.630 & 0.570 & 0.647 & 0.730 & 0.544 & 0.595 & 0.435 & 0.474 & 0.578 \\
  & & SRLP       & 0.406 & 0.578 & 0.587 & 0.631 & 0.566 & 0.736 & 0.057 & \textbf{0.634} & 0.524 \\
  & & SRLP + FS  & 0.468 & 0.605 & 0.651 & 0.743 & 0.605 & 0.736 & 0.251 & {0.504} & 0.571 \\
  \cmidrule{2-12}
  & \multirow{4}{*}{GPT-5.4}
    & CTP        & 0.205 & 0.092 & 0.230 & 0.431 & 0.218 & 0.211 & 0.063 & 0.061 & 0.189 \\
  & & CTP + FS   & 0.311 & 0.133 & 0.304 & 0.621 & 0.330 & 0.309 & 0.193 & 0.066 & 0.283 \\
  & & SRLP       & 0.240 & 0.310 & 0.413 & 0.526 & 0.461 & 0.514 & 0.005 & 0.116 & 0.323 \\
  & & SRLP + FS  & 0.303 & 0.392 & 0.550 & 0.747 & 0.505 & 0.637 & 0.079 & 0.177 & 0.424 \\
  \cmidrule{2-12}
  & \multirow{3}{*}{\refcell{Qwen2.5-7B-Inst$^{\dagger}$}}
    & \refcell{Naive}  & \refcell{0.013} & \refcell{0.006} & \refcell{$-$0.014} & \refcell{0.000} & \refcell{0.030} & \refcell{$-$0.003} & \refcell{0.000} & \refcell{0.019} & \refcell{0.006} \\
  & & \refcell{CTP}    & \refcell{0.000} & \refcell{$-$0.001} & \refcell{0.003} & \refcell{$-$0.056} & \refcell{0.000} & \refcell{0.018} & \refcell{0.042} & \refcell{0.033} & \refcell{0.005} \\
  & & \refcell{SRLP}   & \refcell{0.046} & \refcell{0.075} & \refcell{$-$0.009} & \refcell{0.008} & \refcell{0.152} & \refcell{0.001} & \refcell{$-$0.029} & \refcell{$-$0.043} & \refcell{0.025} \\
\midrule
\multirow{2}{*}{\rotatebox[origin=c]{90}{\textit{SFT}}}
  & Qwen2.5-7B-Inst & SFT & 0.455 & 0.417 & 0.577 & 0.667 & 0.584 & 0.535 & 0.588 & 0.261 & 0.511 {\scriptsize $\pm$0.020} \\
  & Qwen3-4B        & SFT 
  & 0.477 & 0.472 & 0.561 & 0.579 & 0.439 & 0.444 & 0.367 & 0.177 & 0.439 {\scriptsize $\pm$0.055} \\
\midrule
\multirow{2}{*}{\rotatebox[origin=c]{90}{\textit{RL}}}
  & Qwen2.5-7B-Inst & + GRPO$_{\text{PAR}}$ & 0.592 & 0.559 & 0.702 & 0.789 & \textbf{0.741} & 0.731 & \textbf{0.688} & 0.333 & 0.642 {\scriptsize $\pm$0.008} \\
  & Qwen3-4B        & + GRPO$_{\text{PAR}}$ & \textbf{0.683} & \textbf{0.677} & \textbf{0.776} & \textbf{0.814} & 0.733 & \textbf{0.750} & 0.649 & 0.372 & \textbf{0.682} {\scriptsize $\pm$0.060} \\
\bottomrule
\end{tabular}
}
\caption{Prompt-level QWK between target proficiency profiles and AES-predicted profiles on the ASAP/ASAP++ test set. Higher is better. FS: 5-shot prompting. \textbf{Bold} marks the best in each column. $^{\dagger}$Reference rows (in gray) show prompting-only performance of the Qwen2.5-7B-Inst under the zero-shot setting, provided for context.}
\label{tab:prompt-results}
\end{table*}

\begin{table*}[ht!]
\centering
\scalebox{0.8}{
\begin{tabular}{p{0.06\textwidth} p{0.85\textwidth} c}
\toprule
Target & GRPO-generated essay & $\hat{s}$ \\
\midrule

\textbf{1/4} 
& The mood created by the author in the memoir is that he is happy and he 
likes his home. He says that he is grateful to his parents for their love 
and sacrifice. 
\hfill\textit{(33 words, 2 sentences)}
& 1.0 \\
\midrule

\textbf{2/4}
& The mood created by the author in the memoir is that he is proud of his 
parents. He is proud because they gave him a good life. His parents gave 
him a good life by moving to @LOCATION1. They moved to @LOCATION1 so they 
can start a new life. They also gave him a good life because they taught 
him how to cook. His parents taught him how to cook so he can take care 
of himself. 
\hfill\textit{(77 words, 6 sentences)}
& 2.0 \\
\midrule

\textbf{4/4}
& In the memoir ``Narciso Rodriguez'' from \emph{Home: The Blueprints of 
Our Lives}, the author creates a mood. The mood created by the author is 
happiness. The author creates a mood of happiness because he talks about 
how his parents were nice and how he had a lot of friends. The author says 
``My parents always kept their arms and their door open to the many people 
we considered family\ldots'' This shows that his parents were nice and 
kind to everyone. [\textit{Truncated; full essay continues for 224 words with 
3 additional supporting quotations.}]
& 4.0 \\

\bottomrule
\end{tabular}
}
\caption{Same prompt (ASAP Prompt~5, ``mood in \emph{Narciso Rodriguez}''), 
different target Overall scores, generated by GRPO-trained Qwen2.5-7B. 
Length and structural sophistication scale monotonically with target proficiency.
$\hat{s}$ = AES-predicted Overall.}\label{tab:qual-grpo}
\end{table*}

\section{Linguistic Analysis Details}
\label{app:linguistic-metrics}

\paragraph{Cohort construction and aggregation.}
We define low- and high-proficiency cohorts using the prompt-normalized \textit{Overall} score: essays with scores at or below 0.33 constitute the low-target cohort, while those with scores at or above 0.67 constitute the high-target cohort. This yields 247 low-target and 245 high-target essays for each source. All metrics are computed at the essay level and averaged across essays within each source and proficiency cohort. Table~\ref{tab:linguistic_analysis} reports the low-target results.

\paragraph{Error-based metrics.}
We detect \textbf{grammatical}, \textbf{spelling}, and \textbf{casing} errors using LanguageTool\footnote{\url{https://github.com/jxmorris12/language_tool_python}} with the \texttt{en-US} configuration. Grammar errors comprise matches assigned to the \texttt{GRAMMAR} and \texttt{CONFUSED\_WORDS} categories, spelling errors those assigned to \texttt{TYPOS}, and casing errors those assigned to \texttt{CASING}. Each count is normalized by essay length and reported as the number of detected errors per 100 words:
\begin{equation}
\operatorname{Errors/100w}
=
100 \times
\frac{\text{number of detected errors}}
     {\text{number of words}}.
\end{equation}

Before running LanguageTool, we replace dataset anonymization placeholders (e.g., \texttt{@PERSON} and \texttt{@LOCATION}) with fixed, plausible lexical substitutes. We apply the same deterministic replacements to real and generated essays to prevent the placeholders themselves from inflating the detected error counts.

\paragraph{Lexical diversity.}
We measure lexical diversity using the Moving-Average Type--Token Ratio (MATTR; \citealt{covington2010cutting}) with a sliding window of 50 tokens. MATTR is computed over lowercased alphabetic tokens. For essays containing fewer than 50 such tokens, we use all available tokens, making the measure equivalent to conventional type--token ratio for those essays.

\paragraph{Syntactic complexity.}
Following \citet{lu2010automatic} on the automatic analysis of syntactic complexity in learner writing, we measure clausal density and dependency depth. We parse each essay using spaCy's \texttt{en\_core\_web\_sm} model~\citep{honnibal2020spacy}. We operationalize \textbf{clauses per sentence} as a dependency-based proxy: each sentence contributes one main clause, plus one for every token whose dependency label is in \texttt{\{}
\texttt{ccomp},\,
\texttt{xcomp},\,
\texttt{advcl},\,
\texttt{acl},\,
\texttt{relcl},\,
\texttt{csubj},\,
\texttt{csubjpass}
\texttt{\}}.
The resulting clause count is divided by the number of sentences in the essay. We define \textbf{dependency depth} as the maximum number of dependency edges from the sentence root to any token. We compute this maximum separately for each sentence and average it across sentences within an essay.

\definecolor{promptframe}{HTML}{B0B7C3}
\definecolor{promptback}{HTML}{F7F8FA}
\definecolor{titleback}{HTML}{E4E7EC}
\definecolor{essayframe}{HTML}{9FB4D1}
\definecolor{essayback}{HTML}{F2F6FC}
\definecolor{sysframe}{HTML}{C9B79C}
\definecolor{sysback}{HTML}{FBF7F1}

\newtcolorbox{promptbox}[1][]{%
  breakable, enhanced jigsaw,
  colback=promptback, colframe=promptframe,
  boxrule=0.5pt, arc=2pt,
  left=10pt, right=10pt, top=8pt, bottom=8pt,
  fonttitle=\bfseries\small, coltitle=black,
  colbacktitle=titleback,
  attach boxed title to top left={xshift=10pt, yshift=-9pt},
  boxed title style={colframe=promptframe, sharp corners, boxrule=0.4pt},
  fontupper=\small,
  before skip=10pt, after skip=10pt,
  pad at break*=2mm,
  break at=-\baselineskip/0pt,
  #1
}

\newtcolorbox{essaybox}[1][]{%
  breakable, enhanced jigsaw,
  colback=essayback, colframe=essayframe,
  boxrule=0.4pt, arc=1.5pt,
  left=8pt, right=8pt, top=6pt, bottom=6pt,
  fontupper=\footnotesize,
  before skip=6pt, after skip=6pt,
  pad at break*=2mm,
  break at=-\baselineskip/0pt,
  #1
}

\newtcolorbox{sysbox}[1][]{%
  breakable, enhanced jigsaw,
  colback=sysback, colframe=sysframe,
  boxrule=0.5pt, arc=2pt,
  left=10pt, right=10pt, top=8pt, bottom=8pt,
  fonttitle=\bfseries\small, coltitle=black,
  colbacktitle=titleback,
  attach boxed title to top left={xshift=10pt, yshift=-9pt},
  boxed title style={colframe=sysframe, sharp corners, boxrule=0.4pt},
  fontupper=\small,
  before skip=10pt, after skip=10pt,
  pad at break*=2mm,
  break at=-\baselineskip/0pt,
  #1
}

\newcommand{\traitlabel}[1]{\smallskip\noindent\textbf{#1}\par\nobreak\vspace{2pt}}
\newcommand{\sectlabel}[1]{\smallskip\noindent\textsc{\small\textbf{#1}}\par\nobreak\vspace{2pt}}

\clearpage  
\onecolumn
\section{Prompts}
\label{append:prompt}
\begin{sysbox}[title=System Prompt]
You simulate student writing behavior conditioned on overall proficiency and trait-level profile.
\end{sysbox}
\begin{promptbox}[title=Example User Prompt for Contrastive Trait Prompting (CTP)]
You are simulating a student's writing. Your goal is to mirror a specific proficiency level.

\smallskip
The student's writing for this task has been assessed at the following levels:
\begin{itemize}[leftmargin=1.4em, itemsep=0pt, topsep=2pt]
  \item overall: 50 (Range: 0--60)
  \item content: 10 (Range: 2--12)
  \item organization: 10 (Range: 2--12)
  \item word choice: 10 (Range: 2--12)
  \item sentence fluency: 10 (Range: 2--12)
  \item conventions: 10 (Range: 2--12)
  \item voice: 10 (Range: 2--12)
\end{itemize}

\sectlabel{Writing Requirements (Rubric-Specific)}
For each trait, you must strictly adhere to the descriptors below:

\traitlabel{Content:}
A learner with a higher Content level is more likely to provide exceptionally clear, focused, and interesting ideas (i.e., main ideas that stand out and are thoroughly developed with specific, relevant details and strong support), while a learner with a lower level is more likely to provide undeveloped responses (i.e., ideas that are minimally focused, extremely limited, or lack a central purpose). In the event that your commonsense reasoning DIRECTLY conflicts with this hypothesis, use this hypothesis.

\traitlabel{Organization:}
A learner with a higher Organization level is more likely to exhibit strong, clear organization (i.e., connections between ideas and events are clear, logically sequenced, and use a compelling structure that moves the reader through the text easily), while a learner with a lower level is more likely to show weak or no evident organization (i.e., disjointed sequencing where connections between ideas are weak or non-existent). In the event that your commonsense reasoning DIRECTLY conflicts with this hypothesis, use this hypothesis.

\traitlabel{Word Choice:}
A learner with a higher Word Choice level is more likely to use compelling, striking, and varied vocabulary (i.e., a rich, broad range of words thoughtfully placed for impact to clearly support the writer's purpose), while a learner with a lower level is more likely to use extremely limited or colorless vocabulary (i.e., repetitive use of words, incorrect use of words, or vague, imprecise language). In the event that your commonsense reasoning DIRECTLY conflicts with this hypothesis, use this hypothesis.

\traitlabel{Sentence Fluency:}
A learner with a higher Sentence Fluency level is more likely to utilize a wide range of grammatical structures and varied sentence patterns (i.e., demonstrating effective command of language), while a learner with a lower level is more likely to rely on limited language with a lack of variety in sentences or repetitive, grammatically simplistic structures. In the event that your commonsense reasoning DIRECTLY conflicts with this hypothesis, use this hypothesis.

\traitlabel{Conventions:}
A learner with a higher Conventions level is more likely to demonstrate exceptional, consistent control of standard writing conventions (i.e., excellent grammar, usage, spelling, capitalization, and punctuation for the grade level with very few minor errors), while a learner with a lower level is more likely to produce numerous significant errors (i.e., frequent spelling and grammar errors in almost every sentence that impede readability and repeatedly distract the reader). In the event that your commonsense reasoning DIRECTLY conflicts with this hypothesis, use this hypothesis.

\traitlabel{Voice:}
A learner with a higher Voice level is more likely to choose a voice appropriate for the topic, purpose, and audience, demonstrating deep commitment to the topic and a strong sense of ``writing to be read''. Their writing is expressive, engaging, and makes the topic come to life. Conversely, a learner with a lower level is more likely to lack a sense of involvement or commitment, producing writing that is flat, lifeless, stiff, or mechanical. They show a lack of audience awareness with no hint of the writer behind the words. In the event that your commonsense reasoning DIRECTLY conflicts with this hypothesis, use this hypothesis.

\sectlabel{Task}
Write the essay based on the instructions below. Ensure the quality of the writing matches the provided rubric descriptors for each trait (do not exceed them and do not fall below them).

\smallskip
\noindent\textit{Important:}
\begin{itemize}[leftmargin=1.4em, itemsep=0pt, topsep=2pt]
  \item Output ONLY the essay text.
  \item Do NOT mention the scores or the rubric language in the essay itself.
\end{itemize}

\smallskip
\noindent\textbf{Prompt ID:} 8

\sectlabel{Writing Instruction}
We all understand the benefits of laughter. For example, someone once said, ``Laughter is the shortest distance between two people.'' Many other people believe that laughter is an important part of any relationship. Tell a true story in which laughter was one element or part.
\end{promptbox}
\begin{promptbox}[title=Example User Prompt for Score-level Rubric-Lookup Prompting (SRLP)]
You are simulating a student's writing. Your goal is to mirror a specific proficiency level.

\smallskip
The student's writing for this task has been assessed at the following levels:
\begin{itemize}[leftmargin=1.4em, itemsep=0pt, topsep=2pt]
  \item overall: 6 (Range: 2--12)
  \item content: 3 (Range: 1--6)
  \item organization: 3 (Range: 1--6)
  \item word choice: 3 (Range: 1--6)
  \item sentence fluency: 3 (Range: 1--6)
  \item conventions: 3 (Range: 1--6)
\end{itemize}

\sectlabel{Writing Requirements (Rubric-Specific)}
For each trait, you must strictly adhere to the descriptors below:

\traitlabel{Content (score 3):}
The reader can understand the main ideas, although they may be overly broad or simplistic, and the results may not be effective. Supporting detail is often limited, insubstantial, overly general, or occasionally slightly off-topic. The writing is characterized by an easily identifiable purpose and main idea(s); predictable or overly-obvious main ideas; or points that echo observations heard elsewhere; or a close retelling of another work; support that is attempted, but developmental details are often limited, uneven, somewhat off-topic, predictable, or too general (e.g., a list of underdeveloped points); details that may not be well-grounded in credible resources; they may be based on cliches, stereotypes or questionable sources of information; difficulties when moving from general observations to specifics.

\traitlabel{Organization (score 3):}
The essay shows some organization. Its form may not be that of a letter to the editor. Its ideas are not necessarily self-contained.

\traitlabel{Word Choice (score 3):}
Language lacks precision and variety, or may be inappropriate to audience and purpose in places. The writer does not employ a variety of words, producing a sort of ``generic'' paper filled with familiar words and phrases. The writing is characterized by words that work, but that rarely capture the reader's interest; expression that seems mundane and general; slang, if used, does not seem purposeful and is not effective; attempts at colorful language that seem overdone or forced; words that are accurate for the most part, although misused words may occasionally appear; technical language or jargon may be overused or inappropriately used; reliance on cliches and overused expressions; text that is too short to demonstrate variety.

\traitlabel{Sentence Fluency (score 3):}
The writing tends to be mechanical rather than fluid. Occasional awkward constructions may force the reader to slow down or reread. The writing is characterized by some passages that invite fluid oral reading; however, others do not; some variety in sentence structure, length, and beginnings, although the writer falls into repetitive sentence patterns; good control over simple sentence structures, but little control over more complex sentences; fragments, if present, may not be effective; sentences which, although functional, lack energy; lapses in stylistic control; dialogue, if used, may sound stilted or unnatural; text that is too short to demonstrate variety and control.

\traitlabel{Conventions (score 3):}
The writing demonstrates limited control of standard writing conventions (e.g., punctuation, spelling, capitalization, grammar and usage). Errors begin to impede readability. The writing is characterized by some control over basic conventions; the text may be too simple or too short to reveal mastery; end-of-sentence punctuation that is usually correct; however, internal punctuation contains frequent errors; spelling errors that distract the reader; misspelling of common words occurs; capitalization errors; errors in grammar and usage that do not block meaning but do distract the reader; significant need for editing.

\sectlabel{Task}
Write the essay based on the instructions below. Ensure the quality of the writing matches the provided rubric descriptors for each trait (do not exceed them and do not fall below them).

\smallskip
\noindent\textit{Important:}
\begin{itemize}[leftmargin=1.4em, itemsep=0pt, topsep=2pt]
  \item Output ONLY the essay text.
  \item Do NOT mention the scores or the rubric language in the essay itself.
\end{itemize}

\smallskip
\noindent\textbf{Prompt ID:} 1

\sectlabel{Examples}

\begin{essaybox}[title=\textmd{\footnotesize\textbf{Example 1} \;|\; Overall: 12 \;\textbar\; Content: 5 \;\textbar\; Organization: 5 \;\textbar\; Word choice: 5 \;\textbar\; Sentence fluency: 5 \;\textbar\; Conventions: 5}, fonttitle=\normalfont]
``Dear Local Newspaper, I feel that negative effects of computers on people is much greater than the positive effects. Computers have caused a massive amount of obesity in the @LOCATION1 and that number is growing everday. Computers can also be time-consuming and people could spend hours and even days on a computer without doing anything else. Lastly, computers can be dangerouis to many kids and could even end up in death. I feel that xomputers are too harmful for society and should be stopped. Obesity is @CAPS4 of the biggest problems in the @LOCATION1 and is growing bigger everyday. Over @PERCENT1 of @LOCATION1 population is obese and computers have been a major conributer. Put yourself in the mind of a computerf addicy. It is dinner time and you are hungry, but you don't want to lose too much time off of your computer. You hurry down to the nearest fast-food store and by yourself a cheeseburger, fries, and a soda. You then go back to your house and eat the food while on your computer. Not only is the fast-food bad for you, but you dont even give your body a chance to burn @CAPS7 those calories and carbohydrates that you just ate by exercising. Computer addicts do this almost everyday and before they know it, they are. @NUM1 pounds! In addition, people who thier hand-eye coordination can do it in every other ways than just going on a Computers contribute to a large portion of the obese population, but they also can be extremely time-consuming. People on computers waste a huge amount of their timer. These people stay on thier computers for hours amd sometimes they dont even realize it. They say ``@CAPS1, let my check my email. @CAPS1 look, I recieved a @CAPS2! @CAPS1, let me just change my @CAPS3 acconut.'' @CAPS4 thing thing another and @CAPS7 of a madden, @NUM2 or @NUM3 hours have already passed by. People who stay on computers @CAPS7 day waste precious time could be spent with family, exercising, or finding a true love. Time is sao precious in our lives and we must not waste it staying o a computer @CAPS7 day. Not only can they be time consuming, computers can also be dangerous. Kids around the country are being killed, raped, and bulled @CAPS7 to computers. They show thier pictures on chat sites, give away too much information, and enough. They have a critical knocking on their front door. Also, those been who suicide because they were bulled on the @CAPS5. Cyberbullying can be just as harmful. If not more harmful, than vregular bullying. These kids fare such strong mental and emotional harm that they make bad decisions, including suicide, and @NUM4 in school. In addition, there can be explicit content on the @CAPS5, including pictures and video that can harm a child's young mind The danger of computer is so great than they should be @CAPS7 in @CAPS7 computers are too risky for whivh is why they shouldn't be a time consuming. Lastly they can be extremely dangerous to children. Because of these should be banned from @CAPS7 society for the well- using of the @LOCATION1''
\end{essaybox}

\begin{essaybox}[title=\textmd{\footnotesize\textbf{Example 2} \;|\; Overall: 10 \;\textbar\; Content: 5 \;\textbar\; Organization: 5 \;\textbar\; Word choice: 4 \;\textbar\; Sentence fluency: 4 \;\textbar\; Conventions: 4}, fonttitle=\normalfont]
``Dear newspaper, I think computers have a good effect on people. Computers are very helpful to many it gives them an education, and they can interact with many people. Computers can be very helpful to many people. Working with computers can give people extremly well hand-eye coordination. Dr.\ @PERSON4 said, ``In recent surveys and tests, I have noticed a major increase with hand-eye coordination of @PERCENT1 all thanks to computers.'' Computers is also a way that people can learn to type in a recent survey at @ORGANIZATION1, @NUM1-more students know to type, thanks to computers. Computers can be very helpful to business people with creating power points. Instead of making a really big poster board for @CAPS2 @CAPS1 project, I made a power point instead that was so much easier. Computers can be very helpful to many people. Computers are used in many schools to help teach students about new things. For example, students can learn about interesting places that they never knew about. ``@CAPS2 friends is moving to @LOCATION1 and she has no idea what it's like, or where it is. So, we looked it up on the internet and found out so much cool stuff about it,'' says @PERSON5 computers can also teach students about new cultures. At @CAPS3 @ORGANIZATION2, @ORGANIZATION2, @PERCENT2 of the students learned about @NUM2 new cultures in one year, thanks to computers. Not only can computers tell people about interesting places, but interesting people. I never knew that president @PERSON2 had @NUM3 brothers, and gre up in @LOCATION2, did you? Computers can be used in schools to help teach students about new things. Computers are used by many people to interact with another. using computers is an easy way to talk with friends. Using instant messenger is free! Now parents don't have to pay annoying phone bills anymore! In a recent survey done in the town of @CAPS4 @CAPS5, @PERCENT3 of the people there thought that using computers to talk to friends was alot easier. Computers can help people meet new people online. Habe you ever heard of the dating sites? Well, this is what @PERSON1 has to say about it, ``@CAPS6 wonderful! @CAPS2 husband and I met online and we have been together for @NUM4 years.'' @CAPS7 you have a faw away cousin and if you call or write them a letter it's too expensive? Well now you don's have to worry about that anymore because talking to distant relatives online is free! ``@CAPS6 so much easier talking to people online.'' says @PERSON3. Computers can be used to interact with another. Computers have a very good effect on people. They are helpful, can be used to give education, and they can be used to interact with another.''
\end{essaybox}

\begin{essaybox}[title=\textmd{\footnotesize\textbf{Example 3} \;|\; Overall: 7 \;\textbar\; Content: 3 \;\textbar\; Organization: 3 \;\textbar\; Word choice: 3 \;\textbar\; Sentence fluency: 3 \;\textbar\; Conventions: 3}, fonttitle=\normalfont]
``I think we need our computers. If we dont have competer's we wont be able to connect to our family that could live in anther state or country. I no how there is texting now but some people dont even have phones or nothing how can you connect to them you cant call it might not be free with the phone palan you have. If a family member coms and visits you they would probuly email you about it but if there is no computers they can't tell when there coming. Are they gong do show unispectly they might not even be home and if they show up without them knowing there out of luck. And we exersize on our overtime we pick what time and we do it and if we play a sport were enjoying nature some people might have job's that take place outside of if. There are nerd's that are on computers all day long but in sure they enjoy nature I dont know about exersizing but they have a car and every thing they drive and things. Also, about the exersizing every on has to exersize to stay in shape im pretty sure people dont want to be fat there whole life they want to look good and feel good about there self. If it is a holiday or something and people want to send them something to wish then of whetever holiday it is. There not gonna be able to see them in person if they live so far away inless they plan to meet each other but even if they do that there gonna want to email then not call were it costs money and texting only comes with sertain plan's of phones so all there is left is emailing. Sometimes, people will more fron of you that your ollways on the computer if your not dont worry at least you get to email your family that you dont ever seen. These reasons are my opinion that dont get rid of computers and other things.''
\end{essaybox}

\begin{essaybox}[title=\textmd{\footnotesize\textbf{Example 4} \;|\; Overall: 3 \;\textbar\; Content: 1 \;\textbar\; Organization: 1 \;\textbar\; Word choice: 1 \;\textbar\; Sentence fluency: 1 \;\textbar\; Conventions: 1}, fonttitle=\normalfont]
Dear local Newspaper @CAPS1 a take all your computer and given to the people around the world for the can stay in their houses chating with their family and friend. Computers help people around the world to connect with other people computer help kids do their homework and look up staff that happen around the world.
\end{essaybox}

\begin{essaybox}[title=\textmd{\footnotesize\textbf{Example 5} \;|\; Overall: 2 \;\textbar\; Content: 1 \;\textbar\; Organization: 1 \;\textbar\; Word choice: 1 \;\textbar\; Sentence fluency: 1 \;\textbar\; Conventions: 1}, fonttitle=\normalfont]
``Dear readers, I think that its good and bad to use the computer to much''
\end{essaybox}

\sectlabel{Writing Instruction}
More and more people use computers, but not everyone agrees that this benefits society. Those who support advances in technology believe that computers have a positive effect on people. They teach hand-eye coordination, give people the ability to learn about faraway places and people, and even allow people to talk online with other people. Others have different ideas. Some experts are concerned that people are spending too much time on their computers and less time exercising, enjoying nature, and interacting with family and friends.

\smallskip
Write a letter to your local newspaper in which you state your opinion on the effects computers have on people. Persuade the readers to agree with you.
\end{promptbox}


\begin{sysbox}[title=System Prompt for SFT]
You simulate student writing behavior conditioned on trait-level proficiency.
\end{sysbox}

\begin{promptbox}[title=User Prompt for SFT]
You are simulating a student's writing based on trait-level proficiency.

\smallskip
The student receives the following trait scores for this task:

\begin{itemize}[leftmargin=1.4em, itemsep=0pt, topsep=2pt]
  \item Overall: \textit{[Score]} (\textit{[Min]}--\textit{[Max]})
  \item \textit{[Trait 1]}: \textit{[Score]} (\textit{[Min]}--\textit{[Max]})
  \item $\cdots$
  \item \textit{[Trait $N$]}: \textit{[Score]} (\textit{[Min]}--\textit{[Max]})
\end{itemize}

Given the writing instruction below, write an essay that matches this student's trait profile.

\smallskip
\noindent\textit{Important:}
\begin{itemize}[leftmargin=1.4em, itemsep=0pt, topsep=2pt]
  \item Do NOT mention the scores in the essay.
  \item Output ONLY the essay text.
\end{itemize}

\smallskip
\noindent\textbf{Prompt ID:} \textit{[Prompt ID]}

\sectlabel{Writing Instruction}
\textit{[Writing instruction]}
\end{promptbox}

\end{document}